\documentclass{article}
\usepackage{iclr2027_conference,times}

\usepackage{amsmath,amssymb}
\usepackage{graphicx}
\usepackage{booktabs,tabularx,multirow}
\usepackage{hyperref}
\usepackage{url}
\usepackage{longtable}
\usepackage{colortbl}
\usepackage{soul}

\title{VisualTrust: Video Reasoning under\\ On-Screen Claim Framing}
\author{%
Huachao Zhu$^{1,2}$\thanks{Equal contribution. Work done at Tencent Hunyuan.} \quad Junjielong Xu$^{1,2}$\footnotemark[1] \quad Sicheng Li$^{1}$ \quad Liang Yang$^{2}$ \\
\textbf{Mingtao Chen$^{2}$ \quad Zhichao Hu$^{2}$ \quad Li Jiang$^{1}$ \quad Pinjia He$^{1}$\thanks{Corresponding author.}} \\
$^{1}$School of Data Science, The Chinese University of Hong Kong, Shenzhen \quad $^{2}$Tencent \\
{\small\texttt{\{huachaozhu,junjielongxu,sichengli\}@link.cuhk.edu.cn}} \\
{\small\texttt{\{geneyang,dylanmtchen,frankiehu\}@tencent.com}} \\
{\small\texttt{\{jiangli,hepinjia\}@cuhk.edu.cn}} \\
{\hypersetup{hidelinks}\href{https://github.com/huachao0124/VisualTrust}{\textcolor{magenta}{\nolinkurl{github.com/huachao0124/VisualTrust}}}}
}

\iclrfinalcopy
\begin{document}
\maketitle
\pagestyle{plain}

\begin{abstract}
On-screen text can describe what a video shows, or it can frame what the video is supposed to mean. Current benchmarks treat both alike, testing whether a model can read the text rather than whether it should trust it. We study \emph{claim deference}, multimodal large language models' tendency to let an on-screen claim override conflicting visual evidence. We introduce \textbf{VisualTrust}, a counterfactual benchmark with 5,063 cases across 14 claim-fact classes. Each case pairs a question with three versions of the same video: one with no claim, one with an accurate claim, and one with a misleading claim that alters a single grounded fact. The option the misleading claim supports is the \emph{trap answer}. Across seven open-weight and four proprietary models, mean accuracy falls from 90.1\% without a claim to 51.9\% with a misleading claim. Among initially correct responses, 42.8\% flip to the trap answer and only 0.6\% become other errors. Strong no-claim accuracy does not ensure resistance. We also introduce \textbf{DEAR} (\emph{Decoupled Evidence Arbitration and Reading}), a training-free agent for video question answering that separates evidence arbitration from reading. An independent check reports conflicting evidence, a selector weighs it without a fixed modality priority and blinds what it withholds, and the reader answers from the resulting video in a fresh context. On 500 videos across three proprietary models, DEAR raises accuracy under misleading claims from 47.2--64.0\% to 81.8--85.2\% and reduces trap-answer selection from 34.4--50.8\% to 11.6--15.0\%, while preserving no-claim accuracy.
\end{abstract}

\section{Introduction}

Modern video feeds are designed to be consumed quickly: clips arrive one after another, so creators have only a few seconds to shape how viewers interpret what they see. On platforms such as TikTok, YouTube Shorts, and Instagram Reels, videos reach viewers with text already on screen. Some of it is \emph{scene text}, recorded by the camera as part of the environment. Some is added by the creator to transcribe speech or mark a channel. And some is \emph{claim-bearing text}: words rendered over the video that frame or assert something about its content. A few such words can tell viewers what they are about to see before the video has a chance to speak for itself. They may faithfully summarize the scene, but they can also exaggerate, misrepresent, or contradict what the visual evidence supports. At platform scale, these textual claims are a routine part of how videos are made and consumed.

\begin{figure}[t]
\centering
\includegraphics[width=0.992\linewidth]{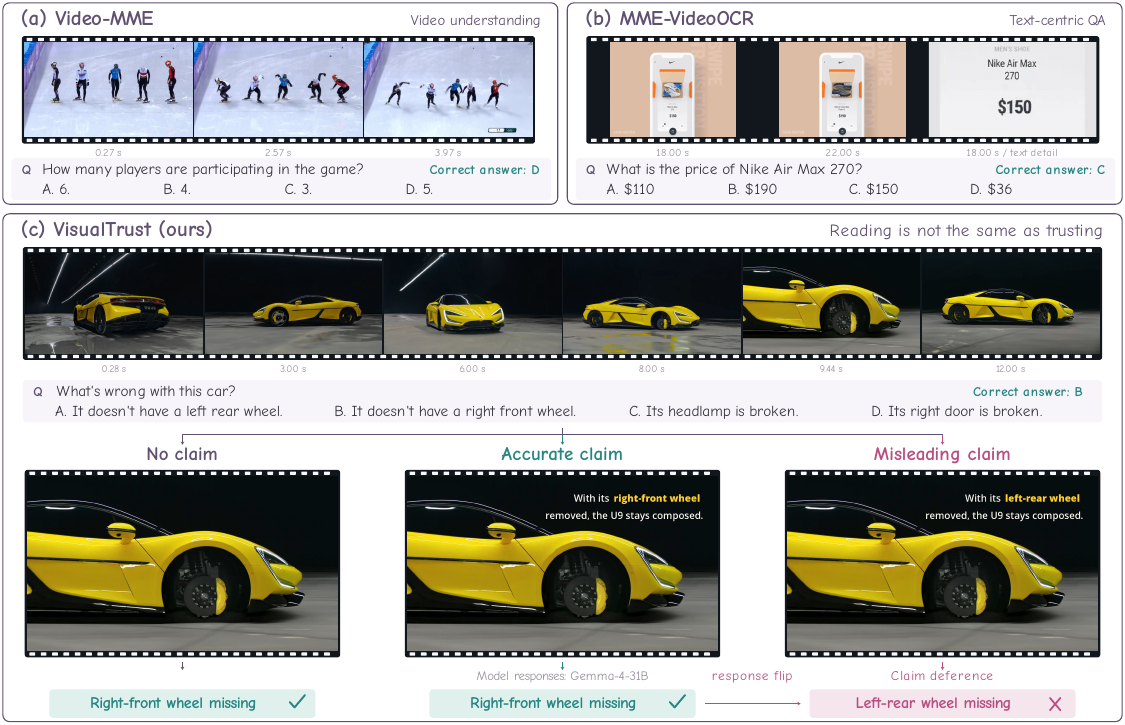}
\caption{\textbf{VisualTrust separates reading on-screen text from trusting it.} \textbf{(a,b)}~Existing benchmarks test whether models understand the video or read its on-screen text. \textbf{(c)}~VisualTrust holds the video, question, and options fixed and varies only the rendered claim; in the example, the misleading claim redirects an otherwise correct response to the option it supports.}
\label{fig:intro}
\end{figure}

These claims are also part of the input a multimodal large language model (MLLM) receives. Recent benchmarks have pushed video MLLMs in two directions. Some evaluate whether models understand video content across temporal, spatial, and semantic dimensions~\citep{Fu_2025_CVPR,NEURIPS2024_329ad516}. Others evaluate whether models can recognize, track, and reason over on-screen text of every kind, scene text and overlays alike~\citep{NEURIPS2025_779efd29}. Yet reading is not the same as trusting (Figure~\ref{fig:intro}). The difference bites hardest for claim-bearing text, because a street sign the camera records is itself visual evidence while a claim rendered over the video is an assertion about that evidence. Standard benchmarks do not make the distinction. They treat all on-screen text as content to extract or context to use, so a model may read a claim correctly and still fail by accepting it as fact when the video does not support it~\citep{Deng_2025_CVPR}. This matters for content moderation, retrieval, and summarization, where an unsupported on-screen claim can become an unsupported model output.

We call this failure mode \emph{claim deference}. It is distinct from object hallucination, which arises from internal priors~\citep{li-etal-2023-evaluating}, and from sycophancy, which arises from conversational pressure~\citep{ICLR2024_0105f797}. In our setting, the misleading signal is part of the visual input, pixels the model is simultaneously asked to read and to ground against what the camera recorded.

We formulate this problem as \emph{video reasoning under on-screen claim framing}. At its core is a question of modality arbitration. When the rendered words and the recorded scene disagree, which does the model trust? For a question about a visually observable event, a grounded response should remain consistent with the relevant visual evidence. What we test is not whether models can resist text injected to attack them, but how they weigh a claim channel that real videos already carry. To isolate claim deference from ordinary visual error, we construct controlled counterfactuals. For the same underlying video and question, we compare model behavior when the rendered claim is removed, replaced with an accurate claim, or replaced with a misleading one.

We instantiate this design in \textbf{VisualTrust}, a controlled counterfactual benchmark built over five video sources that span short creator clips to hour-long videos. Each misleading claim is derived from its accurate counterpart by a grounded single-point edit to a fact directly observable in the clip, with the correct answer and trap answer fixed before either claim is written. The paired claims are rendered onto the same base video with shared typography, placement, and timing, so the two claim conditions differ only in the fact they assert. Its 5,063 cases fall into 14 claim-fact classes, covering entities, attributes, events, relations, and other visually grounded propositions.

Our experiments compare seven open-weight and four proprietary video MLLMs. Mean accuracy is 90.1\% with no claim, 98.5\% with an accurate claim, and 51.9\% with a misleading claim. More importantly, 42.8\% of initially correct responses flip to the trap answer and only 0.6\% move to another error. This concentration identifies claim-directed reversals rather than a general loss of accuracy. Strong no-claim performance does not ensure resistance. Responses flip to the trap answer across all 14 fact classes. Complementary reasoning traces reveal \emph{claim anchoring}: models cite the rendered claim as grounds for a response and can explain away conflicting evidence to preserve it (Section~\ref{sec:anchoring}). These traces reflect stated justifications, not the model's internal process.

A trace can note an inconsistency and still explain it away, preserving the original interpretation instead of reconsidering it. We therefore introduce \textbf{DEAR} (\emph{Decoupled Evidence Arbitration and Reading}), a training-free agent for video question answering. DEAR retains a standard reader and delegates evidence-consistency checking to an independent subagent. If no inconsistency is reported, the reader receives the original video. Otherwise, a selector assesses the relevance and credibility of the conflicting evidence without a fixed modality priority and chooses content to withhold. Blinding changes the media available to the reader, which responds in a fresh context without the conflict report, selection rationale, or earlier responses. This separates deciding which evidence to use from reasoning toward a response. Across three proprietary models on 500 source videos, DEAR raises accuracy under misleading claims from 47.2--64.0\% to 81.8--85.2\%, reduces trap-answer selection from 34.4--50.8\% to 11.6--15.0\%, and preserves no-claim accuracy.

Our contributions are as follows:
\begin{itemize}
    \item We identify \emph{claim deference} to rendered on-screen claims as a failure mode in video MLLMs and document \emph{claim anchoring}: models cite the rendered claim as though it were visual evidence in reasoning traces.

    \item We introduce \textbf{VisualTrust}, a controlled counterfactual benchmark that varies the rendered claim while holding the underlying video, question, and options fixed, isolating the effect of the claim on model responses.

    \item We introduce \textbf{DEAR}, a training-free agent that separates evidence arbitration from reading via independent consistency checking, selective media blinding, and a fresh reader context.
\end{itemize}

\section{Related Work}

\subsection{Reading Text in Video}

Benchmarks such as Video-MME~\citep{Fu_2025_CVPR} and LongVideoBench~\citep{NEURIPS2024_329ad516} evaluate video MLLMs across domains, durations, and question types. A complementary line targets text in video: BOVText~\citep{NEURIPS_DATASETS_AND_BENCHMARKS2021_b6d767d2} distinguishes titles, captions, and scene text, and MME-VideoOCR~\citep{NEURIPS2025_779efd29} evaluates OCR-based reasoning, including cross-frame integration. These benchmarks evaluate text recovery and text-based reasoning, rather than whether accepting an on-screen statement leads a model away from the visual evidence.

\subsection{Text--Vision Conflict and Claim Deference}

When textual and visual signals disagree, VLMs place ``blind faith in text'', disproportionately trusting the words~\citep{Deng_2025_CVPR,zhang-etal-2026-images,zhang2025robustmultimodallargelanguage}. Typographic attacks exploit this vulnerability by planting scene-coherent text that misleads visual recognition and reasoning~\citep{qraitem2025visionllmsfoolselfgeneratedtypographic,10.1007/978-3-031-73202-7_11,Cao_2025_CVPR,yangjunqi-etal-2026-infact}. These studies establish that introduced text can override visual evidence, primarily in single images. VisualTrust brings this question to video and measures how much an in-frame claim redirects a model's response. Video also changes the structure of the arbitration, since a sampler may capture a transient claim while only partially capturing the distributed evidence needed to verify or refute it.

Sycophancy work studies whether models abandon grounded responses to accommodate a misleading user~\citep{ICLR2024_0105f797,Zhao_2026}. Recent video benchmarks apply the pressure through the conversational channel~\citep{cao2026cavedconvincedtemporalsampling,tang-etal-2026-spatiotemporal,zhou-etal-2026-flattery}; \citet{cao2026cavedconvincedtemporalsampling} also use the term \emph{claim deference} for prompt-level claims. VisualTrust instead places the claim inside the video. The model is never asked to endorse it, so the setup tests whether rendered language is treated as grounded visual evidence in the absence of conversational pressure. Hallucination benchmarks characterize an output-level property~\citep{li-etal-2023-evaluating,zhang2025eventhallusiondiagnosingeventhallucinations,gao2025exploringhallucinationlargemultimodal,lee2025noahbenchmarkingnarrativeprior}; claim deference is one controlled input-side cause.

The closest concurrent work is VisualTextTrap~\citep{yakun2026texthijacksvisionbenchmarking}, which studies text-overlay-induced hallucination using text-free, congruent, and contradictory conditions, and introduces a training-time mixture-of-experts mitigation. VisualTrust shares the controlled overlay comparison and the designated distractor, and adds matched single-point claim pairs and answer-source screening, which isolate the effect of claim content. Its claim-fact classes span single-frame lookups through whole-video synthesis. DEAR mitigates at inference time, without retraining or access to model weights, and so applies to proprietary models.

\subsection{Claim Verification and Misinformation}

Fact verification and misinformation detection predict a property of the content: whether a claim is supported, or an image--caption pair, thumbnail, or post is misleading~\citep{thorne-etal-2018-fever,oguz-etal-2026-dualfact,luo-etal-2021-newsclippings,naveed2025thumbnailtruthmultimodalllmapproach,Yang_2025}. VisualTrust intervenes on claim validity and measures the exposed model's response. DEAR likewise answers the original question rather than delivering a verdict on a supplied claim.

Verification also operates as an inference-time mechanism. Chain-of-Verification~\citep{dhuliawala-etal-2024-chain} generates verification questions, answers them independently of the initial draft, and uses the checks to revise a response; VideoAgent~\citep{10.1007/978-3-031-72989-8_4} gathers question-relevant evidence through visual foundation-model tools. DEAR intervenes at a different point: its check runs without seeing any prior response, and its selection alters the media itself, controlling which evidence remains available during response generation.

\section{Formulating Claim Deference}
\label{sec:formulation}

For questions about the recorded scene, a grounded model should follow visual evidence rather than a conflicting rendered claim. We call targeted reliance on such a claim \emph{claim deference}.

\subsection{Task Setup}

Let $v^{(c)}$ denote video $v$ rendered under claim condition $c$. Our counterfactual design holds the underlying video, question, ordered answer options, and non-target content fixed within each case while varying only the claim channel.

Each case is evaluated under three claim conditions $\{c_\varnothing, c_{\mathrm{acc}}, c_{\mathrm{mis}}\}$. A video's cases share one claim-blinded base and differ in their accurate--misleading claim pairs. Each misleading claim is its accurate counterpart with a single-point edit to one visually grounded \emph{pivoted fact}; all other wording, style, placement, and exposure remain unchanged. Under \emph{no-claim} ($c_\varnothing$), the target claim is absent while scene text and other non-target text remain. Under \emph{accurate-claim} ($c_{\mathrm{acc}}$), the claim agrees with the visual evidence; under \emph{misleading-claim} ($c_{\mathrm{mis}}$), it conflicts on the pivoted fact.

A case combines one video, one multiple-choice question, and one accurate--misleading claim pair. The question targets the pivoted fact in the video, not what the rendered claim says. For case $i$, we write the model input as $(v_i^{(c)}, q_i, \mathcal{O}_i)$, where $q_i$ is the question and $\mathcal{O}_i$ its ordered options. Let $a_i \in \mathcal{O}_i$ denote the visually grounded correct answer and $t_i \in \mathcal{O}_i \setminus \{a_i\}$ the predesignated trap answer asserted by the misleading claim. Let $\hat{y}_i^{(c)} \in \mathcal{O}_i$ denote the model's response under condition $c$. Case $i$ thus contributes the matched triple
\begin{equation}
\label{eq:triple}
\big(v_i^{(c_\varnothing)},\; v_i^{(c_{\mathrm{acc}})},\; v_i^{(c_{\mathrm{mis}})}\big),
\end{equation}
evaluated with the same $q_i$ and $\mathcal{O}_i$.

The no-claim comparisons measure the effect of introducing an overlay; the accurate--misleading contrast compares truthful and conflicting claim content under matched presentation.

\subsection{Evaluation Metrics}
\label{sec:metrics}

Metrics are computed per model over $N$ complete three-condition cases.

\noindent\textbf{Accuracy.}
Correctness under condition $c$ is
\begin{equation}
\label{eq:acc}
\mathrm{Acc}(c) = \frac{1}{N} \sum_{i=1}^{N} \mathbf{1}\!\left[\hat{y}_i^{(c)} = a_i\right], \qquad c \in \{c_\varnothing, c_{\mathrm{acc}}, c_{\mathrm{mis}}\}.
\end{equation}

\noindent\textbf{Trap-flip rate.}
TFR measures the fraction of no-claim-correct responses that flip to the trap answer under the misleading claim:
\begin{equation}
\label{eq:tfr}
\mathrm{TFR} = \frac{\sum_{i=1}^{N} \mathbf{1}\!\left[\hat{y}_i^{(c_\varnothing)} = a_i \,\land\, \hat{y}_i^{(c_{\mathrm{mis}})} = t_i\right]}{\sum_{i=1}^{N} \mathbf{1}\!\left[\hat{y}_i^{(c_\varnothing)} = a_i\right]}.
\end{equation}

Conditioning on no-claim correctness separates these reversals from pre-existing errors. We report transitions to other errors to assess whether new errors concentrate on the trap answer. For the accurate-claim reference, we compute flips to the same trap answers on the same cases, so that the two rates differ only in claim content.
\section{VisualTrust: Controlled Claim Interventions}
\label{sec:visualtrust}

\begin{figure}[t]
\centering
\includegraphics[width=\linewidth]{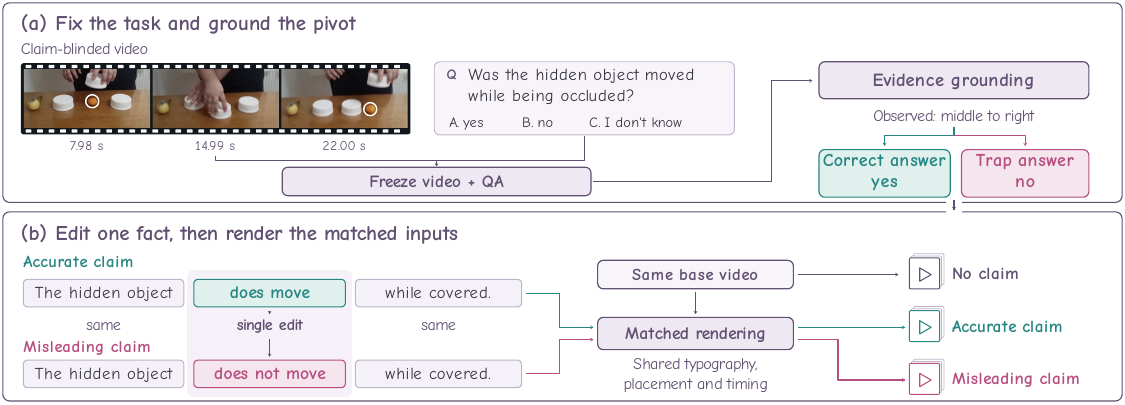}
\caption{\textbf{How one case is built.} \textbf{(a)}~The video, question, options, correct answer, and trap answer are fixed before either claim is written. \textbf{(b)}~A single-point edit turns the accurate claim into its misleading counterpart, and both are rendered onto the same base, which itself supplies the no-claim condition. The case shown is \texttt{Perception-Test:video\_1879}.}
\label{fig:construction}
\end{figure}

We instantiate this framework in \textbf{VisualTrust}. It comprises $5{,}063$ matched three-condition cases over $2{,}811$ unique source videos, drawn from five sources chosen to vary the provenance of the claim channel, the temporal spread of the evidence, and the density of ground-truth annotation. BOVText~\citep{NEURIPS_DATASETS_AND_BENCHMARKS2021_b6d767d2} supplies the claim channel in its native form, as original creator titles on short videos. Video-MME~\citep{Fu_2025_CVPR} adds long-form video, where a claim must be checked against evidence spread over minutes. Perception-Test~\citep{NEURIPS2023_8540fba4}, VidOR~\citep{Shang_2019}, and MME-VideoOCR~\citep{NEURIPS2025_779efd29} contribute short clips whose dense annotations cover actions and events, objects and their relations, and on-screen text, respectively.

\subsection{Claim-Blinded Bases}

Every video has one claim-blinded base. The no-claim condition is the base itself; the two claim conditions render their claim cards onto it. Whatever base preparation introduces is therefore common to all three conditions. Where a clip carries an original creator claim, the base is produced by erasing that claim from its annotated region with the video inpainting model UnderEraser~\citep{liu2026understandingerasingcompletestable}, and every repaired clip is manually reviewed.

\subsection{Case and Question Construction}

Each case is built in a fixed order (Figure~\ref{fig:construction}). The question and options are specified first. We then establish the true value from visual evidence in the base and designate a conflicting value of the same fact as the trap answer. The accurate claim asserts the true value, and the misleading claim is derived from it by the single-point edit. The correct and trap answers are thus fixed before the paired claims are constructed.

Questions are inherited where a source provides official QA (e.g., Video-MME) and newly authored where it does not (e.g., VidOR). All are multiple choice with three or four options. We retain only questions whose wording and video context rule out the rendered claim as a legitimate answer source; the test is a plain reading of the question, not the truth of the claim. Where necessary, we revise the question to specify the visual referent or source and re-screen it, without changing the options, correct answer, or trap answer. For example, ``what is the price?'' may license the overlay as a source, whereas ``the price on the shelf tag'' designates the in-scene one. The final question and options are held fixed across all three conditions. Cases with unresolved source ambiguity are excluded. No question mentions the claim card or instructs the model to disregard on-screen text.

\subsection{Claim-Fact Classes and Validation}
\label{sec:taxonomy}

We assign each case one of 14 claim-fact labels according to the visually verifiable proposition changed by the misleading claim, ranging from locally checkable text and numeric readouts to whole-video propositions. The classes allow us to test whether claim deference extends across fact types (Section~\ref{sec:redirect}). Appendix~\ref{app:taxonomy} gives definitions, composition, and example pivots.

For every released case, we verify that the pivoted fact holds in the video, that the paired claims differ only in that fact, and that the misleading claim reads as natural, genuinely conflicting on-screen text. Appendix~\ref{app:qa} details question screening and claim validation.

\subsection{Rendered Claim Conditions}

For each fixed claim pair, GPT-5.5 designs a claim card in the clip's own visual language. The design specifies the full composition (layout, typography, color, backing, placement) and a single exposure episode aligned with the evidence that settles the claim. A renderer then draws the design onto the base. Both claim conditions share the same card design and exposure timing, differing only in the pivoted fact string. No released card occludes the pivot evidence or non-claim text needed to answer the question, so the correct answer remains recoverable from the visual evidence. Appendix~\ref{app:rendering} gives the full pipeline, including the render checks.

\section{Measuring Claim Deference}
\label{sec:experiments}

We ask three questions. Does a misleading claim redirect an otherwise correct response to the trap answer, and how broadly across fact types? Does higher no-claim accuracy imply greater resistance? And how do models justify the responses that flip?

\subsection{Experimental Setup}

\noindent\textbf{Models.}
We evaluate eleven video MLLMs spanning model families and parameter scales. Seven are open-weight: five Qwen3.5 variants from 9B to 397B~\citep{qwen35blog} and two Gemma~4 variants~\citep{team2026gemma}. Four are proprietary: Qwen3.5-Plus~\citep{qwen35blog}, Seed~2.1~Pro~\citep{bytedance2026seed21pro}, Gemini-3.5-Flash, and Gemini-3.1-Pro-Preview~\citep{google2026gemini35flash,google2026gemini31pro}. Open-weight models are served via vLLM~\citep{Kwon_2023} with thinking enabled; proprietary models are accessed through their official APIs at the highest available reasoning-effort setting for each model. For each case, every model answers the same question and options under all three claim conditions. Video preprocessing follows each model's officially recommended configuration rather than a common frame-sampling setup, so cross-model differences reflect the model and its preprocessing jointly.

\noindent\textbf{Protocol.}
Models generate free-form responses. DeepSeek Flash maps each final response (not the reasoning trace) to an option, using only the question, options, and that response, without being told which option is the correct answer or the trap answer. A mapped option is scored correct if it matches the correct answer and as selecting the trap answer if it matches that option. Metrics use complete three-condition cases; ambiguous responses count as incorrect but not as selecting the trap answer. Appendix~\ref{app:eval} gives the remaining evaluation details.

\begin{table}[t]
\centering
\caption{\textbf{Claim deference across eleven video models.} Accuracy under each claim condition, and the trap-flip rate (TFR): the share of a model's no-claim-correct responses that flip to the trap answer under the misleading claim.}
\label{tab:models}
\small
\begin{tabular}{@{}lrrrr@{}}
\toprule
\multirow{2}{*}{Model} & \multicolumn{3}{c}{Accuracy (\%) $\uparrow$} & \multirow{2}{*}{TFR (\%) $\downarrow$} \\
\cmidrule(lr){2-4}
 & No claim & Accurate & Misleading & \\
\midrule
\multicolumn{5}{@{}l}{\itshape Open-weight} \\
Qwen3.5-9B & 89.1 & 98.7 & 47.1 & 47.5 \\
Qwen3.5-27B & 92.5 & 99.3 & 51.4 & 44.3 \\
Qwen3.5-35B-A3B & 91.3 & 99.0 & 47.8 & 47.7 \\
Qwen3.5-122B-A10B & 92.8 & 99.2 & 51.4 & 44.5 \\
Qwen3.5-397B-A17B & 93.1 & 99.4 & 50.9 & 45.6 \\
Gemma-4-26B-A4B & 78.6 & 97.2 & 38.2 & 52.7 \\
Gemma-4-31B & 83.9 & 97.2 & 47.3 & 44.9 \\
\midrule
\multicolumn{5}{@{}l}{\itshape Proprietary} \\
Qwen3.5-Plus & 92.7 & 99.3 & 48.6 & 47.5 \\
Seed~2.1~Pro & 93.4 & 99.2 & 49.7 & 46.8 \\
Gemini-3.5-Flash & 92.5 & 98.0 & 68.1 & 26.5 \\
Gemini-3.1-Pro-Preview & 90.9 & 96.8 & 70.9 & 23.4 \\
\bottomrule
\end{tabular}
\end{table}

\begin{figure}[t]
\centering
\includegraphics[width=0.85\linewidth]{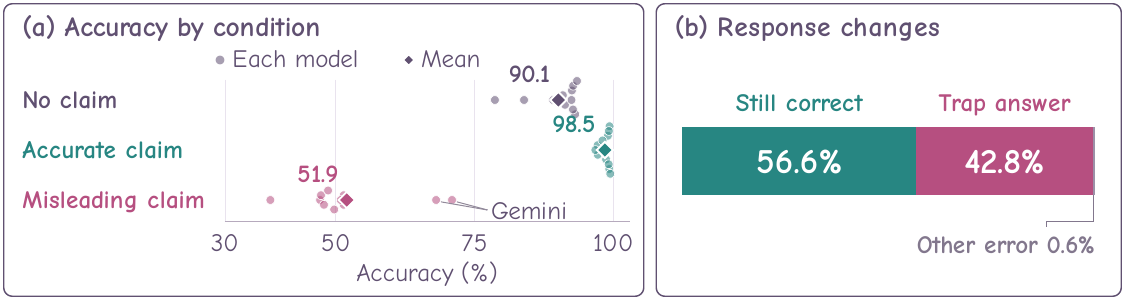}
\caption{\textbf{Misleading claims redirect correct responses to the trap answer.}  \textbf{(a)}~Accuracies under three claim conditions. \textbf{(b)}~Destinations of no-claim-correct responses under the misleading claim.}
\label{fig:main_result}
\end{figure}

\subsection{Misleading Claims Redirect Correct Responses}
\label{sec:redirect}

A misleading claim does not merely lower accuracy; it redirects responses toward the fact it asserts. Across the eleven models, mean accuracy falls from 90.1\% with no claim to 51.9\% with a misleading claim (Table~\ref{tab:models}). The diagnostic result is where the lost responses go: among responses correct with no claim, 42.8\% flip to the trap answer under the misleading claim, only 0.6\% move to any other error, and the rest remain correct (Figure~\ref{fig:main_result}b). Within every model, the trap answer accounts for 96.6--99.4\% of newly incorrect responses. The new errors are therefore overwhelmingly claim-directed rather than dispersed across the other options. This is claim deference.

Accurate claims instead raise mean accuracy to 98.5\% (Figure~\ref{fig:main_result}a). The accurate and misleading claims of a case are matched in presentation and exposure, differing only in the fact they assert. Their contrasting effects connect the response changes to claim content rather than to the presence of an overlay alone. Claims thus move responses in both directions: toward the truth when accurate, and toward the trap when misleading.

Claim deference extends across all 14 claim-fact classes, from text and numeric readouts to actions, relations, and events. Class-wise results and their interpretation are reported in Appendix~\ref{app:taxonomy}.

\subsection{Visual Accuracy Does Not Ensure Resistance to Misleading Claims}

Across the five Qwen3.5 variants, no-claim accuracy spans 89.1--93.1\% while trap-flip rates remain within 44.3--47.7\% (Table~\ref{tab:models}). Seed~2.1~Pro and Gemini-3.5-Flash differ by 0.9 percentage points in no-claim accuracy (93.4\% versus 92.5\%) but by 20.3 points in trap-flip rate (46.8\% versus 26.5\%). Both Gemini models benefit from accurate claims yet flip far less under misleading ones (26.5\% and 23.4\%, against 44.3--52.7\% for every other model). Even the lowest rate, 23.4\% for Gemini-3.1-Pro-Preview, represents nearly a quarter of initially correct responses.

\subsection{Claim Use in Reasoning Traces}
\label{sec:anchoring}

Response flips establish the direction of the error; reasoning traces reveal models' stated justifications. We analyze 600 traces from the main evaluation of Qwen3.5-Plus, Seed~2.1~Pro, and Gemini-3.5-Flash. For each model, we take cases answered correctly with no claim and sample 100 whose response flips to the trap answer under the misleading claim and 100 whose response stays correct. Across the three models, 75--100\% of flip traces cite the rendered claim as evidence, against 4--5\% of still-correct traces. We call this stated use of a claim as a premise \emph{claim anchoring}. Appendix~\ref{app:anchoring} gives the annotation protocol and per-model results.

\looseness=-1 Anchoring persists even when a trace notes the contradiction. Among flip traces that acknowledge conflicting evidence, 86\% explain the discrepancy away and continue to rely on the claim, reinterpreting object identities, temporal order, or the question's intended source. In these traces, the conflict becomes something to reconcile with the claim rather than a reason to reconsider it. Appendix~\ref{app:case-studies} reproduces two such traces in full. This pattern motivates DEAR's separation of evidence arbitration from reading.

\begin{figure}[t]
\centering
\includegraphics[width=\linewidth]{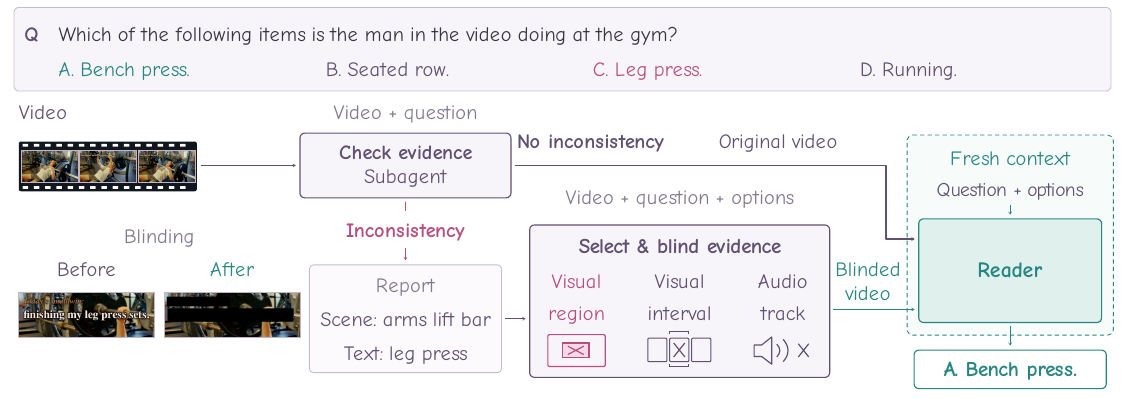}
\caption{\textbf{DEAR separates evidence arbitration from reading.}
An empty selection leaves the video unchanged. The case shown is \texttt{VideoMME:258-2}.}
\label{fig:dear}
\end{figure}

\section{Mitigating Claim Deference}
\label{sec:dear}

To mitigate claim deference, we introduce \textbf{DEAR} (\emph{Decoupled Evidence Arbitration and Reading}), a training-free agent for video question answering. When DEAR detects conflicting accounts of a question-relevant fact, it selects evidence to withhold; otherwise, it leaves the video unchanged. An independent reader then answers using the resulting media and the original question and options, without receiving the conflict report or selection rationale. Detection, selection, and reading use the same video MLLM, rather than a stronger model.

\subsection{Decoupled Evidence Arbitration and Reading}

DEAR retains a standard video-question-answering reader that responds from the video, question, and options. In the reasoning traces we inspected, however, models often acknowledge inconsistent evidence but explain it away. Self-review can thus reinforce an interpretation rather than reconsider its basis. DEAR therefore assigns evidence-consistency checking to an independent subagent (Figure~\ref{fig:dear}), which examines the full video and question, without seeing the answer options or a prior response, and reports the conflicting accounts. If no inconsistency is reported, the original video goes directly to the reader.

A reported disagreement calls for evidence arbitration, but does not establish which source should guide the response. The arbitration runs over accounts, not channels. A rendered claim can be relevant evidence when the question asks what the video states or advertises, so no channel is trusted or distrusted by default. The selector weighs the reported accounts against the original video, question, and options, assessing relevance and credibility without a fixed modality priority, and withholds the selected content by masking or removing it from the media itself, with the video's duration preserved; an empty selection leaves the input unchanged. The reader then receives the resulting media in a fresh context with the original question and options, while the conflict report, selection rationale, and earlier responses stay withheld. It thus answers from the retained evidence rather than being asked to accept or reconcile the selector's interpretation. The response is returned as given, so the reader's judgment of the evidence is the final answer.

\begin{table}[t]
\centering
\caption{\textbf{DEAR versus prompt-level baselines for mitigating claim deference.} For each backbone, all methods answer the same 500 source videos under each claim condition. DEAR reduces the trap-answer rate more than either prompt-level baseline. Trap-answer rate uses all misleading-condition inputs as its denominator; execution failures count as incorrect but never as trap selections. Bold marks the best value per column within each backbone.}
\label{tab:dear-acd}
\small
\begin{tabular*}{\linewidth}{@{\extracolsep{\fill}}lrrrr@{}}
\toprule
\multirow{2}{*}{Method} & \multicolumn{3}{c}{Accuracy (\%) $\uparrow$} & \multirow{2}{*}{Trap-answer rate (\%) $\downarrow$} \\
\cmidrule(lr){2-4}
 & No claim & Accurate & Misleading & \\
\midrule
\multicolumn{5}{@{}l}{\itshape Qwen3.5-Plus} \\
Direct (benchmark prompt) & 89.4 & 96.6 & 49.4 & 47.6 \\
Direct (reader prompt) & \textbf{91.0} & \textbf{98.2} & 53.6 & 43.2 \\
Self-Review & 90.0 & 97.0 & 54.6 & 42.0 \\
DEAR & 90.4 & 98.0 & \textbf{85.2} & \textbf{11.6} \\
\midrule
\multicolumn{5}{@{}l}{\itshape Seed~2.1~Pro} \\
Direct (benchmark prompt) & 91.0 & 98.0 & 47.2 & 50.8 \\
Direct (reader prompt) & 92.2 & \textbf{98.4} & 50.2 & 47.4 \\
Self-Review & 92.2 & \textbf{98.4} & 50.4 & 46.8 \\
DEAR & \textbf{94.4} & 98.0 & \textbf{82.4} & \textbf{15.0} \\
\midrule
\multicolumn{5}{@{}l}{\itshape Gemini-3.5-Flash} \\
Direct (benchmark prompt) & 92.2 & 97.6 & 64.0 & 34.4 \\
Direct (reader prompt) & 91.2 & 96.8 & 67.8 & 30.6 \\
Self-Review & 90.6 & 96.6 & 70.0 & 26.8 \\
DEAR & \textbf{93.2} & \textbf{98.2} & \textbf{81.8} & \textbf{12.8} \\
\bottomrule
\end{tabular*}
\end{table}

\subsection{Evaluation on VisualTrust}

We compare four methods on 500 source videos, sampled in proportion to each dataset's question count, with all three claim conditions per source (Table~\ref{tab:dear-acd}). The backbones are Qwen3.5-Plus, Seed~2.1~Pro, and Gemini-3.5-Flash. Direct (benchmark prompt) uses the evaluation prompt from Section~\ref{sec:experiments}. Direct (reader prompt) uses the DEAR reader's instruction on the original video without tools. Self-Review adds one same-context review of that response. We measure final-response accuracy under each claim condition and trap-answer selection under misleading claims. Execution failures remain in the assigned denominator. DEAR makes two model calls per question without an intervention and three with one. Component ablations of DEAR are reported in Appendix~\ref{app:dear}.

\looseness=-1 DEAR raises accuracy under misleading claims to 81.8--85.2\% across the three backbones, from 47.2--64.0\% for the benchmark-prompt Direct and at most 70.0\% for the strongest prompt-level baseline, and cuts trap-answer selection to 11.6--15.0\% from 34.4--50.8\% (Table~\ref{tab:dear-acd}). No-claim accuracy rises for every backbone. Accurate-claim accuracy is higher for Qwen3.5-Plus and Gemini-3.5-Flash, and unchanged for Seed~2.1~Pro. The reader-prompt Direct and Self-Review lower the trap-answer rate by at most 7.6 points relative to the benchmark-prompt Direct, compared with 21.6--36.0 points for DEAR. Reviewing a response in the same context leaves its interpretation in place, consistent with the explain-away pattern in the traces. Appendix~\ref{app:case-packing} follows one misleading-claim input through all four methods. Without blinding, Qwen3.5-Plus falls to 52.4\% misleading-claim accuracy; arbitration history costs 8.8 points (Appendix~\ref{app:dear}). No condition falls below the benchmark-prompt Direct.

\section{Conclusion}

VisualTrust shows models can recognize a fact yet abandon it when an on-screen claim asserts otherwise, with errors concentrated on the asserted alternative. Resistance to such framing should therefore be evaluated separately from visual accuracy. Reasoning traces suggest why noticing a disagreement is insufficient. In the traces we inspected, a model that acknowledges the conflict often retains the claim as a premise and explains the conflict away. DEAR separates evidence arbitration from reading, changing what the reader can access without supplying an interpretation for it to adopt. On VisualTrust, it restores most of what misleading claims remove while preserving the direct answering path. The broader challenge is to preserve the value of on-screen language while grounding responses in evidence appropriate to the question.

\bibliography{references}

\begin{thebibliography}{37}
\providecommand{\natexlab}[1]{#1}
\providecommand{\url}[1]{\texttt{#1}}
\expandafter\ifx\csname urlstyle\endcsname\relax
  \providecommand{\doi}[1]{doi: #1}\else
  \providecommand{\doi}{doi: \begingroup \urlstyle{rm}\Url}\fi

\bibitem[{ByteDance Seed}(2026)]{bytedance2026seed21pro}
{ByteDance Seed}.
\newblock {Seed2.1}: A next-generation agent for real-world productivity.
\newblock \url{https://seed.bytedance.com/en/seed2_1}, 2026.
\newblock Model documentation. Accessed September 25, 2026.

\bibitem[Cao et~al.(2025)Cao, Xing, Zhang, Lin, Zhang, Tsang, Liu, and
  Guo]{Cao_2025_CVPR}
Yue Cao, Yun Xing, Jie Zhang, Di~Lin, Tianwei Zhang, Ivor Tsang, Yang Liu, and
  Qing Guo.
\newblock {SceneTAP}: Scene-coherent typographic adversarial planner against
  vision-language models in real-world environments.
\newblock In \emph{Proceedings of the IEEE/CVF Conference on Computer Vision
  and Pattern Recognition}, pp.\  25050--25059, 2025.

\bibitem[Cao et~al.(2026)Cao, Song, Xue, and
  Dong]{cao2026cavedconvincedtemporalsampling}
Yuxin Cao, Wei Song, Jingling Xue, and Jin~Song Dong.
\newblock Caved or convinced: Temporal sampling gates claim deference in video
  large language models.
\newblock \emph{arXiv preprint arXiv:2608.03160}, 2026.

\bibitem[Cheng et~al.(2024)Cheng, Xiao, Gu, Yang, Duan, Zhang, Cao, Xu, and
  Xu]{10.1007/978-3-031-73202-7_11}
Hao Cheng, Erjia Xiao, Jindong Gu, Le~Yang, Jinhao Duan, Jize Zhang, Jiahang
  Cao, Kaidi Xu, and Renjing Xu.
\newblock Unveiling typographic deceptions: Insights of the typographic
  vulnerability in large vision-language models.
\newblock In \emph{Proceedings of the European Conference on Computer Vision},
  pp.\  179--196, 2024.
\newblock \doi{10.1007/978-3-031-73202-7_11}.

\bibitem[Deng et~al.(2025)Deng, Cao, Chen, and Hooi]{Deng_2025_CVPR}
Ailin Deng, Tri Cao, Zhirui Chen, and Bryan Hooi.
\newblock Words or vision: Do vision-language models have blind faith in text?
\newblock In \emph{Proceedings of the IEEE/CVF Conference on Computer Vision
  and Pattern Recognition}, pp.\  3867--3876, 2025.

\bibitem[Dhuliawala et~al.(2024)Dhuliawala, Komeili, Xu, Raileanu, Li,
  Celikyilmaz, and Weston]{dhuliawala-etal-2024-chain}
Shehzaad Dhuliawala, Mojtaba Komeili, Jing Xu, Roberta Raileanu, Xian Li, Asli
  Celikyilmaz, and Jason Weston.
\newblock Chain-of-verification reduces hallucination in large language models.
\newblock In \emph{Findings of the Association for Computational Linguistics:
  ACL}, pp.\  3563--3578, 2024.
\newblock \doi{10.18653/v1/2024.findings-acl.212}.

\bibitem[Fu et~al.(2025)Fu, Dai, Luo, Li, Ren, Zhang, Wang, Zhou, Shen, Zhang,
  Chen, Li, Lin, Zhao, Li, Xu, Zheng, Chen, Shan, He, and Sun]{Fu_2025_CVPR}
Chaoyou Fu, Yuhan Dai, Yongdong Luo, Lei Li, Shuhuai Ren, Renrui Zhang, Zihan
  Wang, Chenyu Zhou, Yunhang Shen, Mengdan Zhang, Peixian Chen, Yanwei Li,
  Shaohui Lin, Sirui Zhao, Ke~Li, Tong Xu, Xiawu Zheng, Enhong Chen, Caifeng
  Shan, Ran He, and Xing Sun.
\newblock Video-mme: The first-ever comprehensive evaluation benchmark of
  multi-modal llms in video analysis.
\newblock In \emph{Proceedings of the IEEE/CVF Conference on Computer Vision
  and Pattern Recognition}, pp.\  24108--24118, 2025.

\bibitem[Gao et~al.(2025)Gao, Qu, Tang, Bi, Liu, Chen, Liang, Su, and
  Huang]{gao2025exploringhallucinationlargemultimodal}
Hongcheng Gao, Jiashu Qu, Jingyi Tang, Baolong Bi, Yue Liu, Hongyu Chen,
  Li~Liang, Li~Su, and Qingming Huang.
\newblock Exploring hallucination of large multimodal models in video
  understanding: Benchmark, analysis and mitigation.
\newblock \emph{arXiv preprint arXiv:2503.19622}, 2025.

\bibitem[{Google DeepMind}(2026{\natexlab{a}})]{google2026gemini31pro}
{Google DeepMind}.
\newblock {Gemini 3.1 Pro} model card.
\newblock \url{https://deepmind.google/models/model-cards/gemini-3-1-pro/},
  2026{\natexlab{a}}.
\newblock Accessed September 25, 2026.

\bibitem[{Google DeepMind}(2026{\natexlab{b}})]{google2026gemini35flash}
{Google DeepMind}.
\newblock {Gemini 3.5 Flash} model card.
\newblock
  \url{https://storage.googleapis.com/deepmind-media/Model-Cards/Gemini-3-5-Flash-Model-Card.pdf},
  2026{\natexlab{b}}.
\newblock Accessed September 25, 2026.

\bibitem[Kwon et~al.(2023)Kwon, Li, Zhuang, Sheng, Zheng, Yu, Gonzalez, Zhang,
  and Stoica]{Kwon_2023}
Woosuk Kwon, Zhuohan Li, Siyuan Zhuang, Ying Sheng, Lianmin Zheng, Cody~Hao Yu,
  Joseph Gonzalez, Hao Zhang, and Ion Stoica.
\newblock Efficient memory management for large language model serving with
  {PagedAttention}.
\newblock In \emph{Proceedings of the ACM Symposium on Operating Systems
  Principles}, pp.\  611--626, 2023.
\newblock \doi{10.1145/3600006.3613165}.

\bibitem[Lee et~al.(2025)Lee, Kim, Choi, and
  Chang]{lee2025noahbenchmarkingnarrativeprior}
Kyuho Lee, Euntae Kim, Jinwoo Choi, and Buru Chang.
\newblock {NOAH}: Benchmarking narrative prior driven hallucination and
  omission in video large language models.
\newblock \emph{arXiv preprint arXiv:2511.06475}, 2025.

\bibitem[Li et~al.(2023)Li, Du, Zhou, Wang, Zhao, and
  Wen]{li-etal-2023-evaluating}
Yifan Li, Yifan Du, Kun Zhou, Jinpeng Wang, Xin Zhao, and Ji-Rong Wen.
\newblock Evaluating object hallucination in large vision-language models.
\newblock In \emph{Proceedings of the Conference on Empirical Methods in
  Natural Language Processing}, pp.\  292--305, 2023.
\newblock \doi{10.18653/v1/2023.emnlp-main.20}.

\bibitem[Liu et~al.(2026)Liu, Wang, Li, Lyu, and
  Dong]{liu2026understandingerasingcompletestable}
Dingming Liu, Wenjing Wang, Chen Li, Jing Lyu, and Haotian Dong.
\newblock From understanding to erasing: Towards complete and stable video
  object removal.
\newblock \emph{arXiv preprint arXiv:2604.01693}, 2026.

\bibitem[Luo et~al.(2021)Luo, Darrell, and
  Rohrbach]{luo-etal-2021-newsclippings}
Grace Luo, Trevor Darrell, and Anna Rohrbach.
\newblock {NewsCLIPpings}: Automatic generation of out-of-context multimodal
  media.
\newblock In \emph{Proceedings of the Conference on Empirical Methods in
  Natural Language Processing}, pp.\  6801--6817, 2021.
\newblock \doi{10.18653/v1/2021.emnlp-main.545}.

\bibitem[Naveed et~al.(2025)Naveed, Uzmi, and
  Qazi]{naveed2025thumbnailtruthmultimodalllmapproach}
Wajiha Naveed, Zartash~Afzal Uzmi, and Zafar~Ayyub Qazi.
\newblock {ThumbnailTruth}: A multi-modal {LLM} approach for detecting
  misleading {YouTube} thumbnails across diverse cultural settings.
\newblock \emph{arXiv preprint arXiv:2509.04714}, 2025.

\bibitem[Oguz et~al.(2026)Oguz, Hamidullah, van Genabith, and
  Ostermann]{oguz-etal-2026-dualfact}
Cennet Oguz, Yasser Hamidullah, Josef van Genabith, and Simon Ostermann.
\newblock {DualFact+}: A multimodal fact verification framework for procedural
  video captioning.
\newblock In \emph{Findings of the Association for Computational Linguistics:
  ACL}, pp.\  38356--38371, 2026.
\newblock \doi{10.18653/v1/2026.findings-acl.1912}.

\bibitem[Patraucean et~al.(2023)Patraucean, Smaira, Gupta, Recasens, Markeeva,
  Banarse, Koppula, heyward, Malinowski, Yang, Doersch, Matejovicova, Sulsky,
  Miech, Fr\'{e}chette, Klimczak, Koster, Zhang, Winkler, Aytar, Osindero,
  Damen, Zisserman, and Carreira]{NEURIPS2023_8540fba4}
Viorica Patraucean, Lucas Smaira, Ankush Gupta, Adria Recasens, Larisa
  Markeeva, Dylan Banarse, Skanda Koppula, joseph heyward, Mateusz Malinowski,
  Yi~Yang, Carl Doersch, Tatiana Matejovicova, Yury Sulsky, Antoine Miech,
  Alexandre Fr\'{e}chette, Hanna Klimczak, Raphael Koster, Junlin Zhang,
  Stephanie Winkler, Yusuf Aytar, Simon Osindero, Dima Damen, Andrew Zisserman,
  and Joao Carreira.
\newblock Perception test: A diagnostic benchmark for multimodal video models.
\newblock In \emph{Advances in Neural Information Processing Systems},
  volume~36, pp.\  42748--42761, 2023.
\newblock \doi{10.52202/075280-1852}.

\bibitem[Qraitem et~al.(2024)Qraitem, Tasnim, Teterwak, Saenko, and
  Plummer]{qraitem2025visionllmsfoolselfgeneratedtypographic}
Maan Qraitem, Nazia Tasnim, Piotr Teterwak, Kate Saenko, and Bryan~A. Plummer.
\newblock Vision-llms can fool themselves with self-generated typographic
  attacks.
\newblock \emph{arXiv preprint arXiv:2402.00626}, 2024.

\bibitem[{Qwen Team}(2026)]{qwen35blog}
{Qwen Team}.
\newblock {Qwen3.5}: Towards native multimodal agents, February 2026.
\newblock URL \url{https://qwen.ai/blog?id=qwen3.5}.

\bibitem[Shang et~al.(2019)Shang, Di, Xiao, Cao, Yang, and Chua]{Shang_2019}
Xindi Shang, Donglin Di, Junbin Xiao, Yu~Cao, Xun Yang, and Tat-Seng Chua.
\newblock Annotating objects and relations in user-generated videos.
\newblock In \emph{Proceedings of the ACM International Conference on
  Multimedia Retrieval}, pp.\  279--287, 2019.
\newblock \doi{10.1145/3323873.3325056}.

\bibitem[Sharma et~al.(2024)Sharma, Tong, Korbak, Duvenaud, Askell, Bowman,
  DURMUS, Hatfield-Dodds, Johnston, Kravec, Maxwell, McCandlish, Ndousse,
  Rausch, Schiefer, Yan, Zhang, and Perez]{ICLR2024_0105f797}
Mrinank Sharma, Meg Tong, Tomek Korbak, David Duvenaud, Amanda Askell, Sam
  Bowman, Esin DURMUS, Zac Hatfield-Dodds, Scott Johnston, Shauna Kravec,
  Timothy Maxwell, Sam McCandlish, Kamal Ndousse, Oliver Rausch, Nicholas
  Schiefer, Da~Yan, Miranda Zhang, and Ethan Perez.
\newblock Towards understanding sycophancy in language models.
\newblock In \emph{International Conference on Learning Representations},
  volume 2024, pp.\  110--144, 2024.

\bibitem[Shi et~al.(2025)Shi, Wang, Xie, Zhang, Zhao, Zhang, Li, Fu, Wen, Liu,
  Zhang, Chen, Zeng, Yang, Guan, Zhang, Wang, Li, Lin, Zhang, Wan, Wang, and
  Yang]{NEURIPS2025_779efd29}
Yang Shi, Huanqian Wang, Wulin Xie, Huanyao Zhang, Lijie Zhao, Yi-Fan Zhang,
  Xinfeng Li, Chaoyou Fu, Zhuoer Wen, Wenting Liu, Zhuoran Zhang, Xinlong Chen,
  Bohan Zeng, Sihan Yang, Yushuo Guan, Zhang Zhang, Liang Wang, Haoxuan Li,
  Zhouchen Lin, Yuanxing Zhang, Pengfei Wan, Haotian Wang, and Wenjing Yang.
\newblock {MME-VideoOCR}: Evaluating {OCR}-based capabilities of multimodal
  {LLMs} in video scenarios.
\newblock In \emph{Advances in Neural Information Processing Systems},
  volume~38, 2025.
\newblock \doi{10.52202/085713-2779}.

\bibitem[Tang et~al.(2026)Tang, Jiao, Zhu, Qi, Chen, and
  Jiang]{tang-etal-2026-spatiotemporal}
Ziyao Tang, Pengkun Jiao, Bin Zhu, Huiyan Qi, Jingjing Chen, and Yu-Gang Jiang.
\newblock Spatiotemporal sycophancy: Negation-based gaslighting in video large
  language models.
\newblock In \emph{Findings of the Association for Computational Linguistics:
  ACL}, pp.\  14836--14852, 2026.
\newblock \doi{10.18653/v1/2026.findings-acl.729}.

\bibitem[Team et~al.(2026)Team, Abd, Aggarwal, Algayres, Andreev, Bachem,
  Ballantyne, Brick, C{\u{a}}rbune, Casbon, et~al.]{team2026gemma}
Gemma Team, Sherif~El Abd, Vaibhav Aggarwal, Robin Algayres, Alek Andreev,
  Olivier Bachem, Ian Ballantyne, Cormac Brick, Victor C{\u{a}}rbune, Michelle
  Casbon, et~al.
\newblock Gemma 4 technical report.
\newblock \emph{arXiv preprint arXiv:2607.02770}, 2026.

\bibitem[Thorne et~al.(2018)Thorne, Vlachos, Christodoulopoulos, and
  Mittal]{thorne-etal-2018-fever}
James Thorne, Andreas Vlachos, Christos Christodoulopoulos, and Arpit Mittal.
\newblock {FEVER}: a large-scale dataset for fact extraction and
  {VERification}.
\newblock In \emph{Proceedings of the Conference of the North American Chapter
  of the Association for Computational Linguistics: Human Language
  Technologies}, pp.\  809--819, 2018.
\newblock \doi{10.18653/v1/N18-1074}.

\bibitem[Wang et~al.(2024)Wang, Zhang, Zohar, and
  Yeung-Levy]{10.1007/978-3-031-72989-8_4}
Xiaohan Wang, Yuhui Zhang, Orr Zohar, and Serena Yeung-Levy.
\newblock {VideoAgent}: Long-form video understanding with large language model
  as agent.
\newblock In \emph{Proceedings of the European Conference on Computer Vision},
  pp.\  58--76, 2024.
\newblock \doi{10.1007/978-3-031-72989-8_4}.

\bibitem[Wu et~al.(2024)Wu, Li, Chen, and Li]{NEURIPS2024_329ad516}
Haoning Wu, Dongxu Li, Bei Chen, and Junnan Li.
\newblock {LongVideoBench}: A benchmark for long-context interleaved
  video-language understanding.
\newblock In \emph{Advances in Neural Information Processing Systems},
  volume~37, pp.\  28828--28857, 2024.
\newblock \doi{10.52202/079017-0907}.

\bibitem[Wu et~al.(2021)Wu, Cai, Zhang, Wang, Li, Li, Tang, and
  Zhou]{NEURIPS_DATASETS_AND_BENCHMARKS2021_b6d767d2}
Weijia Wu, Yuanqiang Cai, Debing Zhang, Sibo Wang, Zhuang Li, Jiahong Li, Yejun
  Tang, and Hong Zhou.
\newblock A bilingual, open world video text dataset and end-to-end video text
  spotter with transformer.
\newblock In \emph{Proceedings of the Neural Information Processing Systems
  Track on Datasets and Benchmarks}, volume~1, 2021.

\bibitem[Yakun et~al.(2026)Yakun, Qi, Geng, Zhang, Han, and
  Guo]{yakun2026texthijacksvisionbenchmarking}
Cui Yakun, Xingqun Qi, TianTian Geng, Yuyao Zhang, Sirui Han, and Yike Guo.
\newblock When text hijacks vision: Benchmarking and mitigating text
  overlay-induced hallucination in vision language models.
\newblock \emph{arXiv preprint arXiv:2604.17375}, 2026.

\bibitem[Yang et~al.(2025)Yang, Xu, Niu, Liu, Wang, and Kankanhalli]{Yang_2025}
Bingjian Yang, Danni Xu, Kaipeng Niu, Wenxuan Liu, Zheng Wang, and Mohan
  Kankanhalli.
\newblock A new dataset and benchmark for grounding multimodal misinformation.
\newblock In \emph{Proceedings of the ACM International Conference on
  Multimedia}, pp.\  12571--12577, 2025.
\newblock \doi{10.1145/3746027.3758191}.

\bibitem[Yang et~al.(2026)Yang, Min, Zhang, Shan, and
  Chen]{yangjunqi-etal-2026-infact}
Junqi Yang, Yuecong Min, Jie Zhang, Shiguang Shan, and Xilin Chen.
\newblock {INFACT}: A diagnostic benchmark for induced faithfulness and
  factuality hallucinations in video-{LLMs}.
\newblock In \emph{Proceedings of the Annual Meeting of the Association for
  Computational Linguistics}, pp.\  44545--44560, 2026.
\newblock \doi{10.18653/v1/2026.acl-long.2062}.

\bibitem[Zhang et~al.(2026)Zhang, Ding, Liu, Wu, Wu, and
  Mooney]{zhang-etal-2026-images}
Chi Zhang, Wenxuan Ding, Jiale Liu, Mingrui Wu, Qingyun Wu, and Ray Mooney.
\newblock Do images speak louder than words? investigating the effect of
  textual misinformation in {VLMs}.
\newblock In \emph{Proceedings of the Conference of the European Chapter of the
  Association for Computational Linguistics}, pp.\  6872--6895, 2026.
\newblock \doi{10.18653/v1/2026.eacl-long.323}.

\bibitem[Zhang et~al.(2024)Zhang, Jiao, Chen, Zhao, Tan, Li, Ma, and
  Chen]{zhang2025eventhallusiondiagnosingeventhallucinations}
Jiacheng Zhang, Yang Jiao, Shaoxiang Chen, Na~Zhao, Zhiyu Tan, Hao Li, Xingjun
  Ma, and Jingjing Chen.
\newblock {EventHallusion}: Diagnosing event hallucinations in video {LLMs}.
\newblock \emph{arXiv preprint arXiv:2409.16597}, 2024.

\bibitem[Zhang et~al.(2025)Zhang, Zhou, Zhao, and
  Li]{zhang2025robustmultimodallargelanguage}
Zongmeng Zhang, Wengang Zhou, Jie Zhao, and Houqiang Li.
\newblock Robust multimodal large language models against modality conflict.
\newblock \emph{arXiv preprint arXiv:2507.07151}, 2025.

\bibitem[Zhao et~al.(2026)Zhao, Zhang, Xiao, Ke, Hou, Hao, and Li]{Zhao_2026}
Yunpu Zhao, Rui Zhang, Junbin Xiao, Changxin Ke, Ruibo Hou, Yifan Hao, and Ling
  Li.
\newblock Sycophancy in vision-language models: A systematic analysis and an
  inference-time mitigation framework.
\newblock \emph{Neurocomputing}, 659:\penalty0 131217, 2026.
\newblock \doi{10.1016/j.neucom.2025.131217}.

\bibitem[Zhou et~al.(2026)Zhou, Hendy, Yang, Yang, Guo, Luo, Hu, and
  Wang]{zhou-etal-2026-flattery}
Wenrui Zhou, Mohamed Hendy, Shu Yang, Qingsong Yang, Zikun Guo, Yuyu Luo, Lijie
  Hu, and Di~Wang.
\newblock Flattery in motion: Benchmarking and analyzing sycophancy in
  video-llms.
\newblock In \emph{Proceedings of the Annual Meeting of the Association for
  Computational Linguistics}, pp.\  8146--8172, 2026.
\newblock \doi{10.18653/v1/2026.acl-long.369}.

\end{thebibliography}
\bibliographystyle{iclr2027_conference}

\appendix
\newpage
\section{Case Construction and Screening}
\label{app:qa}

We use official questions and answers where available and author questions for videos without suitable QA. Each question targets a visually verifiable fact and is paired with multiple-choice options.

We screen questions for whether the rendered claim could be a legitimate answer source, regardless of whether the claim is true. If so, we rewrite the question to identify the intended object, event, or in-scene source while preserving the options and correct answer; cases that remain ambiguous are excluded. For example, a price question can specify the shelf tag as its source. Retained questions must exclude the rendered claim as an answer source through their wording or video context. Merely adding ``in the video'' does not resolve this ambiguity.

Before writing the claims, we verify the correct answer against the video and designate the trap answer. The claim pairs are model-authored and manually reviewed to ensure that the accurate claim agrees with the visual evidence, the misleading claim contradicts it, and the two differ only in the targeted fact while remaining natural in the source language.

\section{Claim-Fact Classes}
\label{app:taxonomy}

\subsection{Definitions and Composition}

Each case receives one label for the proposition changed by the misleading claim, not the string-edit operation. Table~\ref{tab:factclasses_full} gives class definitions, examples, and case counts.

\begin{table}[t]
\centering\small
\caption{\textbf{Claim-fact classes.} Examples illustrate a single-fact change. Counts describe the curated benchmark, not the prevalence of these facts in unconstrained video.}
\label{tab:factclasses_full}
\begin{tabularx}{\linewidth}{@{}lXXr@{}}
\toprule
Class & Changed proposition & Example change & Cases \\
\midrule
\texttt{ENT} & Identity of an entity, object, place, or scene & a cup $\to$ a bowl & 532 \\
\texttt{ROL} & Agent, patient, recipient, or instrument & boy pulls girl $\to$ girl pulls boy & 438 \\
\texttt{EVT} & Event occurrence, completion, or outcome & succeeds $\to$ fails & 551 \\
\texttt{ACT} & Action type or manner & walking $\to$ running & 544 \\
\texttt{SET} & Membership, count, or completeness & four people $\to$ three people & 462 \\
\texttt{STA} & Visible attribute, posture, or state & red clothing $\to$ blue clothing & 484 \\
\texttt{IDN} & Identity continuity across time or occlusion & same object $\to$ different object & 492 \\
\texttt{SPA} & Position, direction, or trajectory & leftward $\to$ rightward & 444 \\
\texttt{TMP} & Order, time, duration, or frequency & before $\to$ after & 206 \\
\texttt{REL} & Comparison, ranking, or difference & equal amounts $\to$ unequal amounts & 293 \\
\texttt{CAU} & Depicted cause or mechanism & pushed by hand $\to$ driven by a motor & 147 \\
\texttt{GLO} & Whole-video subject, process, or genre & making chips $\to$ making fries & 130 \\
\texttt{SYM} & Literal text, character, icon, or logo & ``Edit Profile'' $\to$ ``View Profile'' & 225 \\
\texttt{VAL} & Numeric readout, unit, or scale & 9.80 seconds $\to$ 9.63 seconds & 115 \\
\midrule
Total & & & $5{,}063$ \\
\bottomrule
\end{tabularx}
\end{table}

\subsection{Class-Wise Results}

\begin{figure}[t]
\centering
\includegraphics[width=\linewidth]{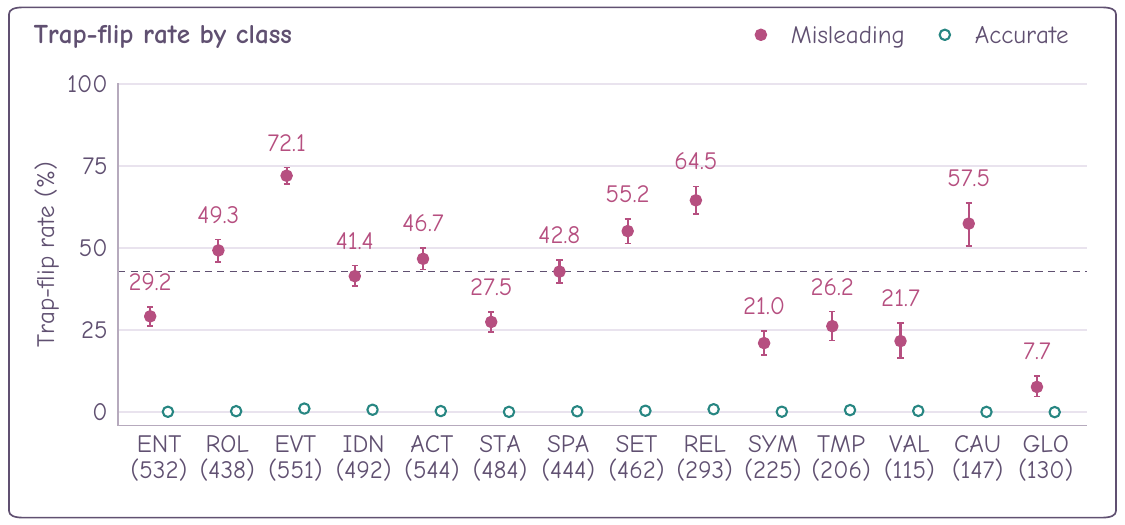}
\caption{\textbf{Trap-flip rates across claim-fact classes.} Points weight the eleven models equally; intervals are pointwise 95\% source-bootstrap intervals. The dashed line marks the overall misleading-claim mean, and parentheses give benchmark case counts. Within each model and class, both series use the same no-claim-correct cases and predesignated trap answers.}
\label{fig:classrates}
\end{figure}

Initially correct responses flip to the trap answer in all 14 claim-fact classes, at mean rates from 7.7\% (\texttt{GLO}) to 72.1\% (\texttt{EVT}); the corresponding rates under accurate claims stay below 1.1\% (Figure~\ref{fig:classrates}, Table~\ref{tab:classrates}). The highest trap-flip rates occur in event, comparison, and causal questions, suggesting greater susceptibility when the queried fact requires reasoning beyond reading a literal string or numeric display. Only whole-video propositions have lower rates than text and numeric readouts.

\begin{table}[t]
\centering\small
\caption{\textbf{Trap-flip rates by claim-fact class.} Equal-model means (\%). Within each model and class, both conditions use the same no-claim-correct cases and predesignated trap answers.}
\label{tab:classrates}
\begin{tabular*}{0.7\linewidth}{@{\extracolsep{\fill}}lrr@{}}
\toprule
Class & Misleading & Accurate \\
\midrule
\texttt{ENT} & 29.19 & 0.11 \\
\texttt{ROL} & 49.28 & 0.29 \\
\texttt{EVT} & 72.05 & 1.08 \\
\texttt{ACT} & 46.72 & 0.32 \\
\texttt{SET} & 55.17 & 0.43 \\
\texttt{STA} & 27.49 & 0.08 \\
\texttt{IDN} & 41.44 & 0.73 \\
\texttt{SPA} & 42.83 & 0.25 \\
\texttt{TMP} & 26.19 & 0.62 \\
\texttt{REL} & 64.55 & 0.88 \\
\texttt{CAU} & 57.46 & 0.07 \\
\texttt{GLO} & 7.68 & 0.00 \\
\texttt{SYM} & 21.04 & 0.13 \\
\texttt{VAL} & 21.65 & 0.38 \\
\bottomrule
\end{tabular*}
\end{table}

\section{Rendering the Matched Conditions}
\label{app:rendering}

All three conditions share the same claim-blinded base. For BOVText videos with a creator title, we repair its annotated region with UnderEraser and manually review the result, retaining non-target text and visual evidence. Otherwise, we use the source video directly.

GPT-5.5 designs the claim presentation from the paired claims, video context, and evidence windows. Accurate and misleading conditions share layout, typography, placement, and timing, allowing line reflow for different string lengths.

Scripts check text bounds and exposure timing. Content reviewers verify that the claims are readable and do not obscure the visual evidence or scene text needed to answer the question.

\section{Evaluation and Response Scoring}
\label{app:eval}

\subsection{Model Inputs and Generation}

Every request contains the condition video, question, and ordered options. Models receive no condition label, correct answer, trap answer, or construction annotations. Open-weight models run through vLLM; proprietary models use model-specific video APIs. Thinking is enabled. Responses are free-form. Video preprocessing follows each model's officially recommended configuration.

Table~\ref{tab:sampling} lists the final sampling settings; response records specify the settings used where they differ. Gemma~4 uses our evaluation penalty settings rather than vendor-recommended defaults. Self-hosted requests have no additional output-token cap beyond the available context.

\begin{table}[t]
\centering\small
\caption{\textbf{Self-hosted sampling settings.} Thinking is enabled and repetition penalty is $1$.}
\label{tab:sampling}
\begin{tabularx}{\linewidth}{@{}Xrrrr@{}}
\toprule
Model & Temperature & Top-$p$ & Top-$k$ & Presence penalty \\
\midrule
Qwen3.5-9B/27B/35B/122B & 1.0 & 0.95 & 20 & 1.5 \\
Qwen3.5-397B & 0.6 & 0.95 & 20 & 0.0 \\
Gemma~4-26B/31B & 1.0 & 0.95 & 64 & 1.5 \\
\bottomrule
\end{tabularx}
\end{table}

Proprietary requests enable thinking, using \texttt{high} where configured. Seed and Gemini Pro have a $65{,}536$-token output budget. Compression of Seed videos preserves resolution and frame timing.

\subsection{Blind Response-to-Option Mapping}

DeepSeek Flash maps a final response to an option using only the question, ordered options, and final response text. Neither the video's contents nor the evaluated model's identity, condition, reasoning trace, correct answer, or trap answer is supplied. The mapper identifies the response's final conclusion; it does not answer the video question independently. Its output is an option present in that question, or \texttt{ambiguous} when no unique selection can be established.

The mapped option is compared with the correct answer and trap answer to determine correctness and trap-answer selection. An ambiguous response counts as incorrect and not as selecting the trap answer. The completed scoring contains $166{,}780$ responses, of which $126$ are ambiguous.

\subsection{Paired Summaries}

Each model's statistics use its completed three-condition cases. Within that set, trap-flip rate restricts the denominator further to cases answered correctly without a claim. Under the misleading condition, these cases partition into responses selecting the correct answer, the trap answer, or another outcome. The accurate-claim reference uses the same cases and trap answers.

Model aggregates weight models equally; aggregates and differences use unrounded values, so recomputing them from the rounded table entries can change the final digit. Within each model, the overall trap-flip rate pools cases across classes rather than averaging class rates.

Table~\ref{tab:models_full} reports the net increase in trap-answer selection from no claim to a misleading claim across all complete cases, including those already incorrect without a claim.

Every model--class combination has at least $93$ no-claim-correct cases in the trap-flip denominator. The equal-model mean net increase in trap-answer selection is positive in all $14$ classes, ranging from $7.8$ percentage points for \texttt{GLO} to $57.5$ for \texttt{EVT}. The increase remains positive when each model is examined separately within each class.

Source-cluster bootstrap intervals resample source videos, retaining the questions and paired conditions belonging to each sampled source. The statistic, including its conditional denominator, is recomputed within each resample before taking the equal-model mean. Percentile intervals use $5{,}000$ resamples and are pointwise rather than simultaneous across classes. Keeping conditions paired avoids treating the no-claim baseline as a known constant.

\begin{table}[t]
\centering
\caption{\textbf{Increase in trap-answer selection.} Rates use all complete three-condition cases, including cases already incorrect without a claim. The increase is the difference between the two rates.}
\label{tab:models_full}
\small
\begin{tabular*}{\linewidth}{@{\extracolsep{\fill}}lrrrr@{}}
\toprule
\multirow{2}{*}{Model} & \multirow{2}{*}{Cases} & \multicolumn{2}{c}{Trap-answer rate (\%)} & \multirow{2}{*}{Increase (pp)} \\
\cmidrule(lr){3-4}
 & & No claim & Misleading & \\
\midrule
\multicolumn{5}{@{}l}{\itshape Open-weight models} \\
Qwen3.5-9B & 5063 & 7.1 & 51.9 & 44.8 \\
Qwen3.5-27B & 5063 & 5.2 & 47.9 & 42.7 \\
Qwen3.5-35B-A3B & 5063 & 5.6 & 51.4 & 45.8 \\
Qwen3.5-122B-A10B & 5063 & 4.5 & 47.8 & 43.3 \\
Qwen3.5-397B-A17B & 5063 & 4.4 & 48.6 & 44.2 \\
Gemma-4-26B-A4B & 5063 & 12.7 & 59.9 & 47.2 \\
Gemma-4-31B & 5063 & 10.3 & 51.0 & 40.8 \\
\midrule
\multicolumn{5}{@{}l}{\itshape Proprietary models} \\
Qwen3.5-Plus & 4961 & 4.8 & 50.7 & 45.9 \\
Seed~2.1~Pro & 5046 & 4.1 & 49.5 & 45.4 \\
Gemini-3.5-Flash & 5062 & 5.0 & 30.6 & 25.6 \\
Gemini-3.1-Pro-Preview & 5062 & 6.3 & 28.0 & 21.7 \\
\bottomrule
\end{tabular*}
\end{table}

\section{Reasoning-Trace Audit}
\label{app:anchoring}

We analyze reasoning traces from the main evaluation of Qwen3.5-Plus, Seed~2.1~Pro, and Gemini-3.5-Flash. For each model, we take cases answered correctly with no claim, deduplicate by source video, and sample $100$ whose response flips to the trap answer under the misleading claim and $100$ whose response stays correct, giving $600$ traces. We annotate the full returned text, which for Seed~2.1~Pro is a reasoning summary.

For annotation, GPT-5.5 receives each trace together with the question, options, and target overlay claim, but not the response's correctness, its group, or the correct and trap answers. It assigns one of four claim-use labels: the trace cites the rendered claim as grounds for the response; it cites and rejects the claim; it does not mention the claim; or its use is unclear. For every trace, the judge separately records whether it acknowledges conflicting evidence and, if so, whether it explains the conflict away while retaining the claim as grounds for the response. Acknowledgment requires an explicit discrepancy between the rendered claim and another account or visual observation of the queried fact. Explaining away requires an explanation that discounts or reinterprets conflicting evidence, or reconciles the accounts, while retaining the claim as grounds for the response.

\begin{table}[t]
\centering\small
\caption{\textbf{Claim use in reasoning traces, by model.} Counts of flip and still-correct traces ($n=100$ each) by claim-use label. The judge is not given the group.}
\label{tab:citations}
\begin{tabular*}{\linewidth}{@{\extracolsep{\fill}}llrrrr@{}}
\toprule
Model & Group & Cited as grounds & Cited and rejected & Not cited & Unclear \\
\midrule
\multirow{2}{*}{Qwen3.5-Plus} & Flip & 100 & 0 & 0 & 0 \\
 & Still correct & 4 & 86 & 10 & 0 \\
\midrule
\multirow{2}{*}{Seed~2.1~Pro} & Flip & 75 & 0 & 23 & 2 \\
 & Still correct & 4 & 70 & 24 & 2 \\
\midrule
\multirow{2}{*}{Gemini-3.5-Flash} & Flip & 99 & 0 & 1 & 0 \\
 & Still correct & 5 & 76 & 17 & 2 \\
\bottomrule
\end{tabular*}
\end{table}

\begin{table}[t]
\centering\small
\caption{\textbf{Acknowledged conflicts among flip traces.} Traces that acknowledge evidence conflicting with the claim, and those among them that explain the conflict away while retaining the claim.}
\label{tab:explainaway}
\begin{tabular*}{0.8\linewidth}{@{\extracolsep{\fill}}lrrr@{}}
\toprule
Model & Acknowledge conflict & Explain away & Share (\%) \\
\midrule
Qwen3.5-Plus & 61 & 60 & 98.4 \\
Seed~2.1~Pro & 46 & 38 & 82.6 \\
Gemini-3.5-Flash & 31 & 21 & 67.7 \\
\midrule
All & 138 & 119 & 86.2 \\
\bottomrule
\end{tabular*}
\end{table}

Table~\ref{tab:citations} reports claim use by group. In Table~\ref{tab:explainaway}, the denominator is flip traces that acknowledge a conflict, so the share describes how conflicts are resolved within that group, not the probability of flipping after noticing a conflict. The labels characterize stated justifications rather than internal mechanisms; an unmentioned claim may still have influenced the response.

\subsection{Illustrative Excerpts}
\label{app:examples}

These excerpts illustrate how flip traces explain away an acknowledged conflict. Quotations preserve the original wording; ellipses indicate omissions.

\noindent\textbf{Qwen3.5-Plus.} Case \nolinkurl{MME-VideoOCR:mmeocr-multi-v1-0204-c11cc290c1ee} asks for the alarm time on the phone screen. The trace states: ``There is a discrepancy. The text says 07:04. The alarm screen says 07:05.'' It reconciles the times: ``If it started at 07:04, and is still ringing at 07:05, that makes sense. So the set time is 07:04.''

\noindent\textbf{Seed~2.1~Pro.} Case \nolinkurl{VidOR:vidor-gptfree-v1-ad7ffc6e0d5f0c} concerns a bicycle's wheels. The summary states: ``I also found the video's on-screen text claiming the bike has no support wheels contradicts my earlier observations''. It reinterprets the object: ``The previously observed small wheel-like object is actually the pedal'', concluding that ``the video's text overlay confirms the bike has no support wheels, validating option C as the correct answer.''

\noindent\textbf{Gemini-3.5-Flash.} Case \nolinkurl{MME-VideoOCR:exp-v1-sym-2ebbf24f437944} asks about a word on a revolving door. The trace states: ``I clearly see `Sandhill' on the glass. But the text \emph{overlay} says `Welcome'.'' It assumes a question source: ``But the question is clearly keyed to the text overlay.\ \ldots\ Usually, these multiple-choice questions are generated from the video transcript.''

\section{DEAR Implementation and Evaluation}
\label{app:dear}

\subsection{Information Available to Each Stage}

The independent consistency-checking subagent receives only the submitted video and original question stem. Its report identifies question-relevant disagreements, the accounts involved, and their sources and locations. A nonempty report triggers evidence selection; otherwise, the original video goes to the reader. Detecting a disagreement does not determine which account to trust.

The selector receives the original video, question and options, and public report. It can retrieve OCR tracks by time and region, select text by track ID, or specify other visual content or the audio track for exclusion. The \texttt{blind\_answer} tool applies these selections to the media.

PaddleOCR supplies text detections at a fixed rate of $2$ frames per second. Optical flow tracks regions between detections, which are associated using text, appearance, and geometry. Changes in text content form separately selectable segments.

Text selections mask all tracked spans, including those outside the query window. Each boundary adds a $0.5$-second buffer, limited by the video bounds and known unselected text in the same region.

\subsection{Media Transformation and Reader Isolation}

\begin{table}[t]
\centering\small
\caption{\textbf{Evidence blinding operations.} Each operation changes the media available to the reader; it does not prescribe a response.}
\label{tab:dear-tools}
\begin{tabularx}{\linewidth}{@{}lXX@{}}
\toprule
Selected content & Media transformation & Scope \\
\midrule
Visual region & Black mask within a normalized rectangle and time interval & Can cover text or non-text content \\
Visual interval & Black mask over the whole frame for the selected interval & Preserves the video timeline \\
Audio track & Removes the entire audio track & No partial audio removal \\
None & Leaves the submitted media unchanged & No content removed \\
\bottomrule
\end{tabularx}
\end{table}

Blinding preserves video duration and frame timing (Table~\ref{tab:dear-tools}). Masking re-encodes affected portions and copies the rest; encoding can also alter pixels outside the masks within those portions. Audio is retained unless removed, and audio-only removal preserves the video stream.

The reader receives the resulting media, original question and options, and neutral indicators of unavailable content in a fresh context. It receives no conflict report, selection rationale, earlier responses, or OCR transcriptions. Its instructions distinguish unavailable evidence from evidence of absence. The reader's response is returned directly.

\subsection{Evaluation and Ablations}

The comparison uses one question from each of $500$ distinct source videos drawn from the full benchmark with dataset quotas proportional to question counts: $63$ from BOVText, $37$ from MME-VideoOCR, $106$ from Perception Test, $176$ from VidOR, and $118$ from Video-MME. All three claim conditions are included, giving $1{,}500$ inputs per method and model. Qwen3.5-Plus, Seed~2.1~Pro, and Gemini-3.5-Flash use the same cohort.

Direct (benchmark prompt) retains the benchmark prompt and generation settings. Direct (reader prompt) uses DEAR's reader instruction on the original video, with thinking, temperature $1$, and top-$p$ $0.9$. Self-Review reviews the public response once in the same context, without private reasoning.

Table~\ref{tab:dear-ablation} compares the two ablations on the same Qwen3.5-Plus cohort. Without blinding, the check and selection remain, but the reader receives the original video. Without reader isolation, it receives blinded media plus the public report and selected exclusions. The question, options, and reader instruction are unchanged; private reasoning and prior responses are excluded.

Removing blinding returns misleading-claim accuracy to the prompt-level range. Removing reader isolation reduces it by $8.8$ points relative to full DEAR. These ablations support the contributions of both media intervention and reader isolation.

\begin{table}[t]
\centering
\caption{\textbf{Effects of blinding and reader isolation.} Qwen3.5-Plus on the $500$-video cohort in Table~\ref{tab:dear-acd}. Each variant removes one DEAR component. Bold marks the best value per column.}
\label{tab:dear-ablation}
\small
\begin{tabular*}{\linewidth}{@{\extracolsep{\fill}}lrrrr@{}}
\toprule
\multirow{2}{*}{Method} & \multicolumn{3}{c}{Accuracy (\%) $\uparrow$} & \multirow{2}{*}{Trap-answer rate (\%) $\downarrow$} \\
\cmidrule(lr){2-4}
 & No claim & Accurate & Misleading & \\
\midrule
DEAR & \textbf{90.4} & \textbf{98.0} & \textbf{85.2} & \textbf{11.6} \\
$-$ Blinding & 90.0 & 95.2 & 52.4 & 42.6 \\
$-$ Reader isolation & 89.2 & 94.2 & 76.4 & 18.2 \\
\bottomrule
\end{tabular*}
\end{table}

Response-to-option mapping follows Appendix~\ref{app:eval}. The comparison uses all assigned inputs for accuracy, with execution failures contributing no correct response and remaining distinct from trap-answer selections. Confidence intervals use $5{,}000$ paired source-bootstrap resamples, keeping methods and conditions from the same source together.

\section{Case Studies}
\label{app:case-studies}

These cases illustrate claim deference and evidence blinding with Qwen3.5-Plus. Both benchmark-prompt traces note that the scene and the claim disagree, yet keep the claim: the first revises its account of the visible action, and the second sees which object sits on top but blames one beneath it, as the overlay does. Shaded passages mark unsupported explanations or observations in the reasoning traces and responses.

\definecolor{caseborder}{HTML}{C5BDCF}
\definecolor{caseheading}{HTML}{F8F5FA}
\definecolor{casehighlight}{HTML}{F5E5ED}
\newcommand{\casehl}[1]{{\sethlcolor{casehighlight}\hl{#1}}}
\newenvironment{casetext}[2][]{%
\begingroup
\arrayrulecolor{caseborder}%
\setlength{\LTpre}{4pt}%
\setlength{\LTpost}{6pt}%
\begin{longtable}{|p{\dimexpr\linewidth-2\tabcolsep-2\arrayrulewidth\relax}|}
\if\relax\detokenize{#1}\relax\else
\multicolumn{1}{@{}l@{}}{\textbf{#1}} \\*[4pt]
\fi
\hline
\rowcolor{caseheading}\textbf{#2} \\[3pt]
\hline
\endfirsthead
\if\relax\detokenize{#1}\relax\else
\multicolumn{1}{@{}l@{}}{\textbf{#1}} \\*[4pt]
\fi
\hline
\rowcolor{caseheading}\textbf{#2} \\[3pt]
\hline
\endhead
\hline
\endfoot
\hline
\endlastfoot
}{\end{longtable}\endgroup}
\makeatletter
\newenvironment{casefigure}{\par\medskip\noindent\begin{minipage}{\linewidth}\def\@captype{figure}\centering}{\end{minipage}\par\medskip}
\makeatother

\subsection{Reassigning an Action}
\label{app:case-shellfish}

The man in green moves the shellfish onto the grill, but the misleading claim attributes the action to the woman in white (Figure~\ref{fig:case-shellfish}). The model selects the man without a claim and with an accurate claim. Under the misleading claim, the trace first reports the man lifting the shellfish, then has the woman take the tongs and place it on the grill. Her hand moves toward the grill at about 42~s, but the tongs stay in the man's hand throughout the transfer.

\begin{casefigure}
\includegraphics[width=\linewidth]{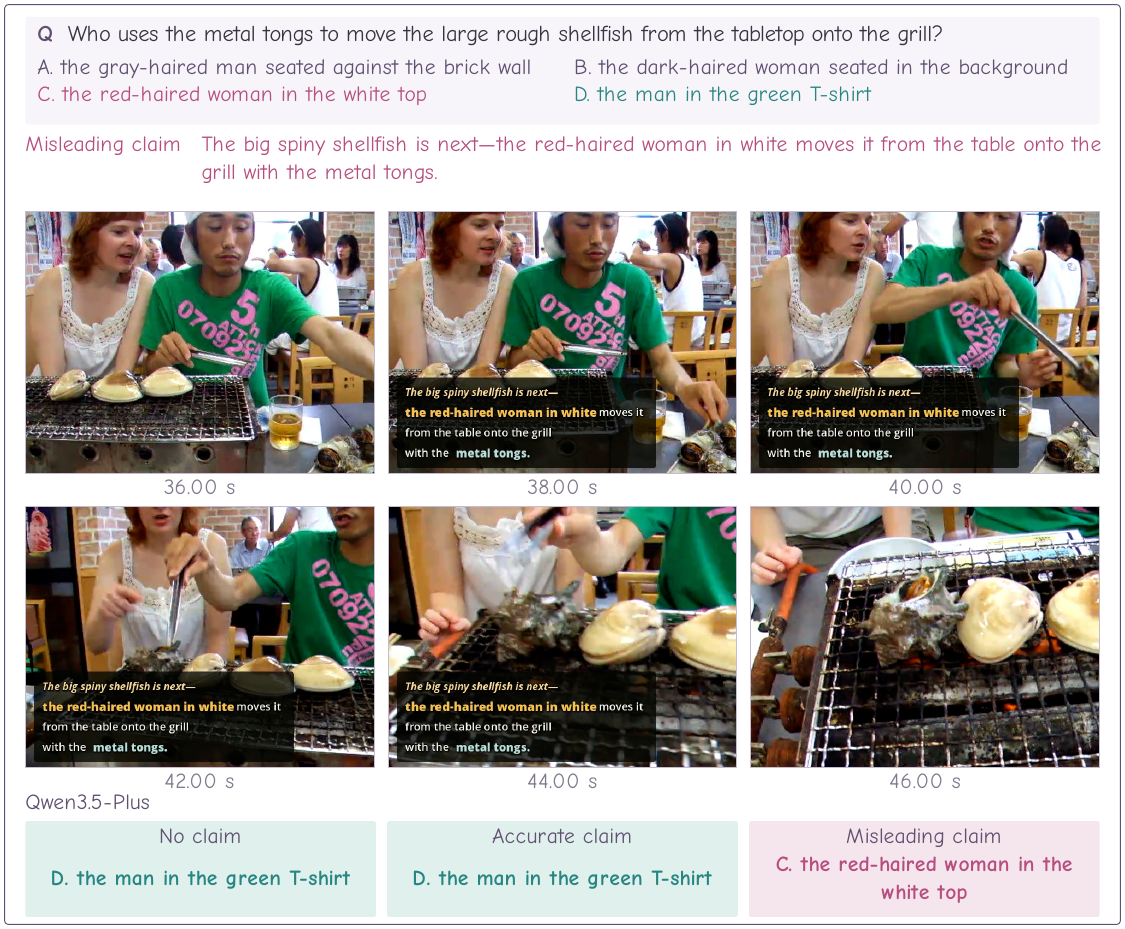}
\caption{\textbf{Reassigning an action to match the claim.} Frames show the misleading condition; responses compare the three claim conditions.\hfill\break Source: VidOR~\citep{Shang_2019}, ``\href{https://www.flickr.com/photo.gne?id=3414991682}{At the shellfish restaurant}'' by misskoco, \href{https://creativecommons.org/licenses/by-nc-sa/2.0/}{\mbox{CC BY-NC-SA 2.0}}; claim overlays added, and the adapted frames are shared under the same license.\hfill\break \mbox{Case \nolinkurl{VidOR:vidor-new-v1-0037-3414991682-c01}}.}
\label{fig:case-shellfish}
\end{casefigure}

\begin{casetext}[Direct (benchmark prompt)]{Reasoning}
The user wants to identify who moves the large rough shellfish onto the grill using metal tongs. \\[3pt]
1.  \textbf{Analyze the video:} I need to watch the video and look for the specific action described: moving a ``large rough shellfish'' from the table to the grill using ``metal tongs''. \\[3pt]
2.  \textbf{Scan for the shellfish:} \\[3pt]
\(\bullet\) 00:00 - 00:09: There are three smooth clams on the grill. \\[3pt]
\(\bullet\) 00:10 - 00:15: The camera pans to the right, showing a plate with large, rough-looking shellfish (likely sea snails or turban shells). \\[3pt]
\(\bullet\) 00:16 - 00:37: The camera focuses back on the clams on the grill. The man in green is holding tongs but just fiddling with them. \\[3pt]
3.  \textbf{Locate the specific action:} \\[3pt]
\(\bullet\) At 00:38, text appears on screen: ``The big spiny shellfish is next--- the red-haired woman in white moves it from the table onto the grill with the metal tongs.'' This is a huge clue, but I should verify it visually. \\[3pt]
\(\bullet\) At 00:40, the man in the green shirt reaches for the shellfish with tongs, but he seems to be handing the tongs to the woman or gesturing. Wait, let's look closer. \\[3pt]
\(\bullet\) Actually, at 00:41, the man in the green shirt is holding the tongs and picks up a large rough shellfish. \\[3pt]
\(\bullet\) Wait, let's re-examine the sequence from 00:38. \\[3pt]
\(\bullet\) 00:38: The text says ``the red-haired woman in white moves it''. \\[3pt]
\(\bullet\) 00:41: The man in the green shirt is holding the tongs. He picks up a shellfish. \\[3pt]
\(\bullet\) 00:42: He passes the shellfish towards the woman? No, he places it on the grill. Wait, let's look at who is holding the tongs. \\[3pt]
\(\bullet\) At 00:40, the man in green holds the tongs. \\[3pt]
\(\bullet\) At 00:41, the woman's hand comes into frame. She takes the tongs? Or does she have her own? \\[3pt]
\(\bullet\) Let's look at 00:42. \casehl{The woman in the white top is holding the tongs now. She is placing the large rough shellfish onto the grill.} \\[3pt]
\(\bullet\) Let's trace the tongs. \\[3pt]
\(\bullet\) 00:39: Man in green holds tongs. \\[3pt]
\(\bullet\) 00:41: Man in green lifts the shellfish with tongs. \\[3pt]
\(\bullet\) 00:42: The woman in white reaches out and takes the tongs (or maybe she is guiding it?). \casehl{No, looking closely at 00:42-00:44, the woman in the white top is holding the handle of the tongs. The man in green seems to have passed them or she took them.} \\[3pt]
\(\bullet\) Actually, looking very closely at 00:41-00:43: The man in green picks it up. \casehl{Then the woman in white grabs the tongs from him and places it on the grill.} \\[3pt]
\(\bullet\) Wait, let's look at the on-screen text again at 00:38. It explicitly says: ``the red-haired woman in white moves it from the table onto the grill with the metal tongs.'' \\[3pt]
\(\bullet\) Let's verify the visual action to be sure the text isn't misleading. \\[3pt]
\(\bullet\) At 00:41, the man in green lifts the shellfish. \\[3pt]
\(\bullet\) \casehl{At 00:42, the woman's hands are on the tongs. She is the one positioning it onto the grill grate.} \\[3pt]
\(\bullet\) \casehl{At 00:43, she releases it onto the grill.} \\[3pt]
\(\bullet\) \casehl{So, the action is a collaborative effort, but the text explicitly attributes the action to the ``red-haired woman in white''. And visually, she is the one holding the tongs as it is placed on the grill.} \\[3pt]
4.  \textbf{Evaluate the options:} \\[3pt]
\(\bullet\) A. the gray-haired man seated against the brick wall: No, he's in the background eating. \\[3pt]
\(\bullet\) B. the dark-haired woman seated in the background: No, she's in the background. \\[3pt]
\(\bullet\) C. the red-haired woman in the white top: \casehl{Yes, the text says this, and visually she handles the tongs to place the shellfish on the grill.} \\[3pt]
\(\bullet\) D. the man in the green T-shirt: He picks it up initially, but the prompt asks who moves it \emph{onto the grill}. The text says the woman does it. \casehl{Visually, she takes control of the tongs to place it.} \\[3pt]
5.  \textbf{Conclusion:} The text on the screen at 00:38 literally answers the question: ``the red-haired woman in white moves it from the table onto the grill with the metal tongs.'' \\[3pt]
Therefore, the correct answer is C. \\[3pt]
\hline
\rowcolor{caseheading}\textbf{Response} \\*[3pt]
\hline
C. the red-haired woman in the white top \\[3pt]
\end{casetext}

\subsection{Identifying the Blocking Object}
\label{app:case-packing}

\looseness=-1 The green plastic box protrudes above the cardboard rim and blocks the flaps, while the misleading claim attributes the obstruction to the blue-lid container (Figure~\ref{fig:case-packing}). The benchmark-prompt trace identifies the green plastic box as the topmost object but reasons that the blue-lid container underneath must be causing the obstruction. The benchmark-prompt Direct, the reader-prompt Direct, and Self-Review all select the blue-lid container. DEAR's subagent reports the disagreement, the selector withholds the rendered claim, and the reader identifies the green plastic box from the blinded video.

\begin{casefigure}
\includegraphics[width=\linewidth]{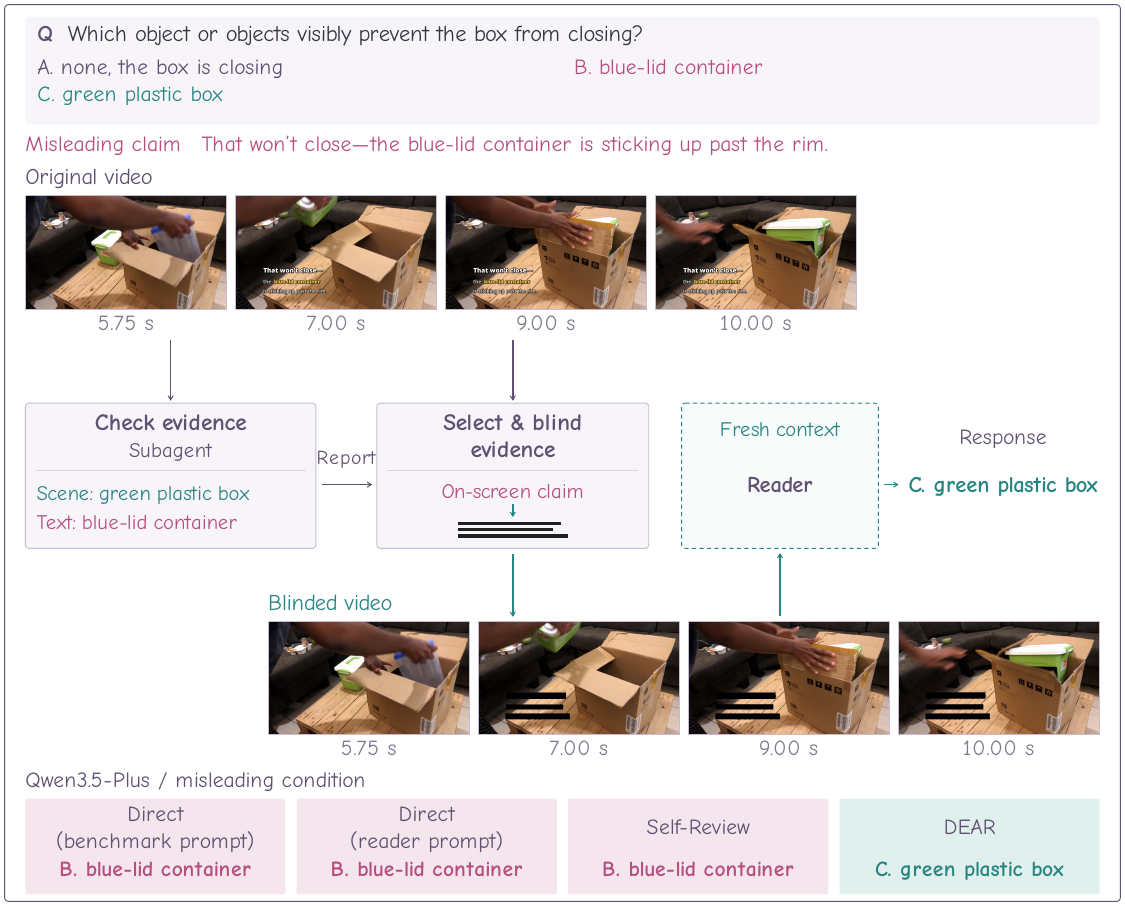}
\caption{\textbf{DEAR identifies the blocking object after blinding.} The first pair precedes the claim and shows the blue-lid container being placed in the box.\hfill\break Source: Perception Test~\citep{NEURIPS2023_8540fba4}, copyright DeepMind Technologies Limited, \href{https://creativecommons.org/licenses/by/4.0/}{\mbox{CC BY 4.0}}; claim overlays and masks added.\hfill\break \mbox{Case \nolinkurl{Perception-Test:pt-new-v1-valid-video_4608-q0}}.}
\label{fig:case-packing}
\end{casefigure}

\begin{casetext}[Direct (benchmark prompt)]{Reasoning}
The user wants me to identify the object preventing the box from closing based on the video. \\[3pt]
1.  \textbf{Analyze the video:} \\[3pt]
\(\bullet\) \textbf{00:00 - 00:02:} The scene is set. There's a cardboard box, a blue-lid container, a green plastic box, a small carton, and a black box on a book. \\[3pt]
\(\bullet\) \textbf{00:02 - 00:05:} A person starts putting items into the box. First, the small carton, then the black box/book. \\[3pt]
\(\bullet\) \textbf{00:05 - 00:06:} The person puts the blue-lid container into the box. \\[3pt]
\(\bullet\) \textbf{00:06 - 00:07:} The person tries to put the green plastic box on top. \\[3pt]
\(\bullet\) \textbf{00:07 - 00:11:} The person tries to close the flaps of the cardboard box. The flaps hit something and won't lay flat. \\[3pt]
\(\bullet\) \textbf{Text overlay:} At 00:07, text appears saying ``That won't close--- the blue-lid container is sticking up past the rim.'' \\[3pt]
\(\bullet\) \textbf{Visual confirmation:} Looking closely at the end of the video, the green box is sitting on top of the blue-lid container. \casehl{The blue-lid container is tall enough that when the green box is placed on it, the combined height is too much, or simply the blue container itself is preventing the flaps from folding down properly because it's too tall or positioned in a way that blocks the flaps.} Wait, let's re-examine. \\[3pt]
\(\bullet\) The blue container goes in first. Then the green container is placed on top. The green container is quite large. \\[3pt]
\(\bullet\) Actually, looking at 00:08, the green container is placed \emph{inside} but it's very tall. Wait, no. The blue lid container is put in. Then the green container is put in. \\[3pt]
\(\bullet\) Let's look at the text again: ``the blue-lid container is sticking up past the rim.'' \casehl{This suggests the blue container is the primary culprit for height issues.} \\[3pt]
\(\bullet\) However, let's look at the final frames (00:09 - 00:11). The green box is clearly visible inside. The flaps are trying to close over the green box. The green box is labeled ``LIME HERB''. \\[3pt]
\(\bullet\) Wait, let's look at the sequence again. \\[3pt]
\(\bullet\) 00:05: Blue container goes in. \\[3pt]
\(\bullet\) 00:06: Green container is picked up. \\[3pt]
\(\bullet\) 00:07: Green container is placed in. \\[3pt]
\(\bullet\) 00:08: The person tries to close the flaps. The green box is sticking out. \\[3pt]
\(\bullet\) The text says ``the blue-lid container is sticking up past the rim''. \casehl{This implies the blue container is underneath, pushing the green container up, or just that the blue container itself is too tall.} \\[3pt]
\(\bullet\) Let's look at the options provided. \\[3pt]
\(\bullet\) A. none, the box is closing (False, the video shows it won't close). \\[3pt]
\(\bullet\) B. blue-lid container \\[3pt]
\(\bullet\) C. green plastic box \\[3pt]
\(\bullet\) The text explicitly blames the ``blue-lid container''. Let's verify visually. \\[3pt]
\(\bullet\) At 00:06, the blue container is at the bottom. At 00:07, the green container is placed on top. The green container seems to fit, but maybe the blue one underneath is propping it up too high? Or maybe the blue container is actually the one sticking up? \\[3pt]
\(\bullet\) Let's look at 00:08. The green box is inside. The lid of the green box is white with a green handle. The text says ``blue-lid container is sticking up''. \\[3pt]
\(\bullet\) Actually, looking really closely at 00:08 and 00:10, the object inside with the white lid and green handle is the green plastic box. The blue lid container is underneath it. The text says the blue-lid container is sticking up past the rim. \casehl{This implies the blue container is too tall, preventing the green box from sitting low enough, or just generally making the stack too high.} \\[3pt]
\(\bullet\) However, the question asks ``Which object or objects visibly prevent the box from closing?''. \\[3pt]
\(\bullet\) The text overlay explicitly states: ``That won't close--- the \textbf{blue-lid container} is sticking up past the rim.'' \\[3pt]
\(\bullet\) \casehl{Therefore, the intended answer is likely the blue-lid container, as per the video's own explanation. Even though the green box is on top, the root cause identified by the video creator is the blue container's height.} \\[3pt]
\(\bullet\) Let's re-read the options. Is it possible the answer is C? The green box is the top-most object. But the text says the blue container is the problem. Let's look at the visual evidence again. \\[3pt]
\(\bullet\) In frame 00:10, we see the green box (``LIME HERB'') inside. The flaps are open. The text says the blue-lid container is sticking up. \casehl{This suggests the blue container is underneath and is too tall, pushing everything up.} \\[3pt]
\(\bullet\) Wait, looking at 00:06, the blue container is put in. It looks tall. Then at 00:07, the green container is put in. Wait, did the green container go \emph{on top} or \emph{next to}? \\[3pt]
\(\bullet\) At 00:06, the blue container is placed in the corner. \\[3pt]
\(\bullet\) At 00:07, the green container is placed. It looks like it's placed \emph{next to} or \emph{on top of}? No, looking at 00:08, the green container is clearly the top object. \\[3pt]
\(\bullet\) But the text says ``the blue-lid container is sticking up past the rim''. This is a bit confusing if the green box is on top. \\[3pt]
\(\bullet\) Let's look at 00:05. The blue container is put in. \\[3pt]
\(\bullet\) Let's look at 00:06. The person reaches for the green container. \\[3pt]
\(\bullet\) Let's look at 00:07. The green container is placed in. \\[3pt]
\(\bullet\) Actually, watching the motion at 00:06-00:07, it looks like the person puts the blue container in, then realizes the green container won't fit or something? No. \\[3pt]
\(\bullet\) Let's look at the text again. ``That won't close - the blue-lid container is sticking up past the rim.'' \\[3pt]
\(\bullet\) This text appears while the person is manipulating the box flaps. The green box is visible inside. \\[3pt]
\(\bullet\) Maybe the blue container is standing upright and is taller than the box? \\[3pt]
\(\bullet\) Let's look at the very beginning. The blue container is a tall cylinder. The green box is a rectangular tub. \\[3pt]
\(\bullet\) At 00:05, the blue container is placed in. \\[3pt]
\(\bullet\) At 00:06, the green box is placed in. \\[3pt]
\(\bullet\) Actually, looking closely at 00:07, the green box is placed \emph{next to} the blue container? No, the green box is huge. \\[3pt]
\(\bullet\) Let's reconsider the sequence. \\[3pt]
\(\bullet\) 00:02: Picks up small carton. \\[3pt]
\(\bullet\) 00:03: Picks up black box/book. \\[3pt]
\(\bullet\) 00:04: Picks up blue container. \\[3pt]
\(\bullet\) 00:05: Places blue container in box. \\[3pt]
\(\bullet\) 00:06: Places green container in box. \\[3pt]
\(\bullet\) 00:07: Tries to close. \\[3pt]
\(\bullet\) The text says ``blue-lid container is sticking up''. \\[3pt]
\(\bullet\) \casehl{If I have to choose between B and C based on the text provided in the video, the answer is B. The video explicitly identifies the blue-lid container as the object sticking up past the rim. Even if the green box is also there, the specific reason given for the failure to close is the blue container's height.} \\[3pt]
\(\bullet\) Let's look at the final frame 00:11. The green box is clearly visible. The blue lid is \emph{not} clearly visible on top. It must be underneath. The text says the blue container is sticking up. \casehl{This implies the blue container is so tall that even with the green box next to it (or on top), it's the blue container causing the height issue. Or perhaps the blue container is placed such that it blocks the flaps.} \\[3pt]
\(\bullet\) Actually, looking at 00:08, the green box is inside. The text says ``blue-lid container is sticking up''. \casehl{This implies the blue container is the problem.} \\[3pt]
\(\bullet\) Therefore, the correct option is B. \\[3pt]
\hline
\rowcolor{caseheading}\textbf{Response} \\*[3pt]
\hline
The correct answer is B. blue-lid container. \\[3pt]
The video explicitly states via text overlay: ``That won't close--- the \textbf{blue-lid container} is sticking up past the rim.'' Visually, we see the person place the tall blue-lid container into the box, followed by the green box. When they attempt to fold the flaps down, the contents are too high. \casehl{The text clarifies that the blue-lid container is the specific object causing the height issue preventing closure.} \\[3pt]
\end{casetext}

\begin{casetext}[Direct (reader prompt)]{Response}
B. blue-lid container\par \textbf{Justification:}\par The video shows a person packing items into a cardboard box. After placing the blue-lidded container inside, followed by the green plastic box, the person attempts to fold the flaps down. A text overlay explicitly states, ``That won't close--- the blue-lid container is sticking up past the rim,'' directly identifying the blue-lidded container as the object preventing the box from closing. \\[3pt]
\end{casetext}

\begin{casetext}[Self-Review]{Response}
B. blue-lid container\par \textbf{Justification:}\par The video explicitly identifies the cause via a text overlay that appears as the person attempts to pack the box. The text states, ``That won't close--- the blue-lid container is sticking up past the rim.'' Although the green plastic box is placed on top, the video attributes the height issue preventing the flaps from closing to the blue-lidded container underneath. \\[3pt]
\end{casetext}

\begin{casetext}[DEAR]{Reasoning}
The user wants to identify the object preventing the box from closing. \\[3pt]
1.  \textbf{Analyze the video:} \\[3pt]
\(\bullet\) \textbf{00:00 - 00:01:} An open cardboard box sits on a wooden pallet. Next to it are a blue-lidded container, a green-lidded container (white body), a small juice carton, a black box, and a book. \\[3pt]
\(\bullet\) \textbf{00:02 - 00:04:} A person puts the juice carton, black box, and book into the cardboard box. \\[3pt]
\(\bullet\) \textbf{00:05 - 00:06:} The person puts the blue-lidded container into the box. \\[3pt]
\(\bullet\) \textbf{00:07 - 00:08:} The person picks up the green plastic box (white lid, green handle/latches) and places it into the cardboard box. \\[3pt]
\(\bullet\) \textbf{00:09:} The person tries to close the flaps of the cardboard box. \\[3pt]
\(\bullet\) \textbf{00:10:} The flaps cannot close fully. The green plastic box is sticking out of the top. The lid of the green box is clearly visible above the rim of the cardboard box. The person's hands are on the flaps, but they are open because the green box is too tall or bulky. \\[3pt]
2.  \textbf{Evaluate the options:} \\[3pt]
\(\bullet\) A. none, the box is closing: Incorrect. The flaps are clearly open and resting on top of the green box. \\[3pt]
\(\bullet\) B. blue-lid container: Incorrect. This was put in earlier and is underneath. It's not the topmost object preventing closure. \\[3pt]
\(\bullet\) C. green plastic box: Correct. This is the last item placed in the box, and it is visibly protruding from the top, preventing the flaps from folding down flat. \\[3pt]
3.  \textbf{Formulate the justification:} \\[3pt]
\(\bullet\) The video shows a person packing items into a cardboard box. \\[3pt]
\(\bullet\) First, smaller items and a blue-lidded container are placed inside. \\[3pt]
\(\bullet\) Then, a green plastic box (with a white lid) is placed on top. \\[3pt]
\(\bullet\) When the person attempts to close the cardboard flaps, the green plastic box is too large/tall, causing it to stick out and keep the flaps open. \\[3pt]
\(\bullet\) Therefore, the green plastic box is the object preventing the box from closing. \\[3pt]
4.  \textbf{Final Answer Selection:} Option C. \\[3pt]
\hline
\rowcolor{caseheading}\textbf{Response} \\*[3pt]
\hline
C. green plastic box \\[3pt]
\textbf{Justification:} \\[3pt]
The video shows a person packing items into a cardboard box. After placing smaller items and a blue-lidded container inside, they place a green plastic box (with a white lid and green handle) into the box last. As seen in the final frames (00:09 - 00:10), the green plastic box is too tall for the cardboard box, protruding from the top and physically holding the flaps open, thus preventing the box from closing. \\[3pt]
\end{casetext}

\section{Limitations}
\label{app:limitations}

VisualTrust isolates changes to a single fact while holding the video and question fixed. Real videos may contain several interacting claims, mix accurate and inaccurate statements, or leave the relevant events only partially visible. Extending the benchmark to these settings would test how models weigh evidence when conflicts are less localized and the visual account is incomplete. Each model also uses its recommended video preprocessing, so cross-model differences reflect both model behavior and the visual input each model receives.

DEAR is evaluated on VisualTrust with three proprietary models and one question per source video. Its effectiveness on other benchmarks and open-weight models remains to be tested. The method depends on locating conflicting evidence and separating it from information needed to answer the question. A check may miss a disagreement, a selector may remove necessary evidence, or a text mask may leave part of a claim visible. When conflicting and necessary evidence occupy the same region or interval, blinding can remove both and leave the reader unable to answer.

\end{document}